\documentclass{article}
\usepackage{iclr2027_conference,times}
\usepackage{authblk}
\usepackage[T1]{fontenc}
\usepackage{amsmath,amsfonts,bm}

\def\eqref#1{equation~\ref{#1}}
\def\1{\bm{1}}

\DeclareMathAlphabet{\mathsfit}{\encodingdefault}{\sfdefault}{m}{sl}
\SetMathAlphabet{\mathsfit}{bold}{\encodingdefault}{\sfdefault}{bx}{n}

\newcommand{\arxivshuffleglyph}[1]{%
  \scalebox{#1}{\setlength{\unitlength}{1em}%
    \begin{picture}(0.889,0.5)%
      \put(0.061,0){\rule{0.767em}{0.04em}}%
      \put(0.061,0){\rule{0.04em}{0.5em}}%
      \put(0.424,0){\rule{0.04em}{0.5em}}%
      \put(0.788,0){\rule{0.04em}{0.5em}}%
    \end{picture}}}
\newcommand{\shuffle}{\mathbin{\mathchoice
  {\arxivshuffleglyph{1}}{\arxivshuffleglyph{1}}%
  {\arxivshuffleglyph{0.7}}{\arxivshuffleglyph{0.5}}}}
\usepackage{amsmath,amssymb}
\usepackage{booktabs}
\usepackage{array}
\usepackage{graphicx}
\usepackage{float}
\usepackage{placeins}
\usepackage{algorithm}
\usepackage{algpseudocode}
\usepackage{multirow}
\usepackage{microtype}
\usepackage{xcolor}
\usepackage{enumitem}
\usepackage{hyperref}
\hypersetup{
  colorlinks=false,
  pdfborder={0 0 1},
  citebordercolor={0 1 0},
  pdfauthor={Jinglin He, Siyang Jiang, Lixing He, Guoliang Xing, Hongkai Chen},
  pdftitle={Word-Level Text Unmixing via Evidence-Preserving Ownership Routing with Language Models}
}
\usepackage{url}
\usepackage{caption}
\usepackage{xspace}
\usepackage[subtle]{savetrees}
\definecolor{PromptFrame}{HTML}{3F3F3F}
\definecolor{PromptSystem}{HTML}{174A8B}
\definecolor{PromptSystemBg}{HTML}{EAF2FC}
\definecolor{PromptUser}{HTML}{176B5B}
\definecolor{PromptUserBg}{HTML}{EAF7F3}
\definecolor{PromptOutput}{HTML}{9A5200}
\definecolor{PromptOutputBg}{HTML}{FFF3E3}
\definecolor{PromptVariable}{HTML}{2675B8}
\newcommand{\promptvar}[1]{\textcolor{PromptVariable}{\bfseries\{#1\}}}
\newcommand{\promptheader}[1]{%
  \noindent\colorbox{PromptFrame}{%
    \parbox{\dimexpr\linewidth-2\fboxsep\relax}{%
      \vspace{2pt}\hspace{7pt}\color{white}\sffamily\bfseries #1\vspace{2pt}}}}
\newcommand{\promptband}[3]{%
  \noindent\colorbox{#1}{%
    \parbox{\dimexpr\linewidth-2\fboxsep\relax}{%
      \vspace{2pt}\hspace{6pt}\textcolor{#2}{\sffamily\bfseries #3}\vspace{2pt}}}}


\graphicspath{{figures/}}

\newcommand{\problem}{word-level text unmixing\xspace}
\newcommand{\method}{\textsc{EPoR}\xspace}
\newcommand{\benchmark}{\textsc{UnMixBench}\xspace}

\newcommand{\MPWER}{\textsc{MP-WER}\xspace}
\newif\ifrealserializationresults
\realserializationresultstrue
\newif\ifturnmarkerresults
\turnmarkerresultstrue
\newif\ifsimilarityresults
\similarityresultstrue
\newcommand{\ConfirmFreeWERMean}{0.6696}
\newcommand{\ConfirmFreeWERSD}{0.0608}
\newcommand{\ConfirmFreeEvidenceMean}{0.0845}

\newcommand{\ConfirmFreeParseSuccess}{0.6972}
\newcommand{\ConfirmFreeParseSuccessSD}{0.1147}
\newcommand{\ConfirmFreeConditionalWER}{0.5272}
\newcommand{\ConfirmFreeConditionalWERSD}{0.0108}
\newcommand{\ConfirmFreeConditionalEvidence}{0.1239}
\newcommand{\ConfirmFreeConditionalEvidenceSD}{0.0259}

\newcommand{\ConfirmGlobalWERMean}{0.5368}
\newcommand{\ConfirmGlobalWERSD}{0.0192}
\newcommand{\ConfirmGlobalTokenAccMean}{0.6658}

\newcommand{\ConfirmGlobalSwitchFOneMean}{0.6847}

\newcommand{\ConfirmGlobalEvidenceMean}{1.0000}

\newcommand{\ConfirmEmaWERMean}{0.5165}
\newcommand{\ConfirmEmaWERSD}{0.0140}
\newcommand{\ConfirmEmaTokenAccMean}{0.6703}

\newcommand{\ConfirmEmaSwitchFOneMean}{0.7390}

\newcommand{\ConfirmEmaEvidenceMean}{1.0000}

\newcommand{\ConfirmCrfWERMean}{0.4799}
\newcommand{\ConfirmCrfWERSD}{0.0063}
\newcommand{\ConfirmCrfTokenAccMean}{0.6917}

\newcommand{\ConfirmCrfSwitchFOneMean}{0.7001}

\newcommand{\ConfirmCrfEvidenceMean}{1.0000}

\newcommand{\ConfirmProjectionWERMean}{0.4788}
\newcommand{\ConfirmProjectionWERSD}{0.0140}
\newcommand{\ConfirmProjectionTokenAccMean}{0.6904}

\newcommand{\ConfirmProjectionSwitchFOneMean}{0.7695}

\newcommand{\ConfirmProjectionEvidenceMean}{1.0000}

\newcommand{\ConfirmDirectRgsWERMean}{0.3757}
\newcommand{\ConfirmDirectRgsWERSD}{0.0056}
\newcommand{\ConfirmDirectRgsTokenAccMean}{0.7457}

\newcommand{\ConfirmDirectRgsSwitchFOneMean}{0.8514}

\newcommand{\ConfirmDirectRgsEvidenceMean}{1.0000}

\newcommand{\CrossWikiCrfWERMean}{0.3583}
\newcommand{\CrossWikiCrfWERSD}{0.0081}

\newcommand{\CrossWikiDirectWERMean}{0.2506}
\newcommand{\CrossWikiDirectWERSD}{0.0136}

\newcommand{\CrossWikiDirectMinusCrfWER}{-0.1077}
\newcommand{\CrossWikiDirectMinusCrfWERCI}{[-0.1174, -0.0982]}

\newcommand{\CrossAgCrfWERMean}{0.6187}
\newcommand{\CrossAgCrfWERSD}{0.0220}

\newcommand{\CrossAgDirectWERMean}{0.7322}
\newcommand{\CrossAgDirectWERSD}{0.0228}

\newcommand{\CrossAgDirectMinusCrfWER}{0.1135}
\newcommand{\CrossAgDirectMinusCrfWERCI}{[0.1056, 0.1213]}

\newcommand{\CrossAmiCrfWERMean}{0.6483}
\newcommand{\CrossAmiCrfWERSD}{0.0103}

\newcommand{\CrossAmiDirectWERMean}{0.6320}
\newcommand{\CrossAmiDirectWERSD}{0.0086}

\newcommand{\CrossAmiDirectMinusCrfWER}{-0.0163}
\newcommand{\CrossAmiDirectMinusCrfWERCI}{[-0.0257, -0.0062]}

\newcommand{\CrossIcsiCrfWERMean}{0.6581}
\newcommand{\CrossIcsiCrfWERSD}{0.0036}

\newcommand{\CrossIcsiDirectWERMean}{0.6283}
\newcommand{\CrossIcsiDirectWERSD}{0.0057}

\newcommand{\CrossIcsiDirectMinusCrfWER}{-0.0297}
\newcommand{\CrossIcsiDirectMinusCrfWERCI}{[-0.0457, -0.0113]}

\newcommand{\MechanismBeamOneSelectedWER}{0.3916}
\newcommand{\MechanismBeamThirtyTwoSelectedWER}{0.3737}
\newcommand{\MechanismBeamThirtyTwoOracleWER}{0.2031}
\newcommand{\MechanismBeamThirtyTwoWERRegret}{0.1706}

\newcommand{\PIAuditDirectWERMean}{0.3757}
\newcommand{\PIAuditDirectWERSD}{0.0056}
\newcommand{\PIAuditDirectTokenAccMean}{0.7457}
\newcommand{\PIAuditDirectTokenAccSD}{0.0039}
\newcommand{\PIAuditDirectSwitchFOneMean}{0.8514}
\newcommand{\PIAuditDirectSwitchFOneSD}{0.0022}
\newcommand{\PIAuditDirectEvidenceMean}{1.0000}

\newcommand{\PIAuditDirectStructuralMean}{1.0000}

\newcommand{\PIAuditPIWERMean}{0.3782}
\newcommand{\PIAuditPIWERSD}{0.0012}
\newcommand{\PIAuditPITokenAccMean}{0.7438}
\newcommand{\PIAuditPITokenAccSD}{0.0037}
\newcommand{\PIAuditPISwitchFOneMean}{0.8521}
\newcommand{\PIAuditPISwitchFOneSD}{0.0023}
\newcommand{\PIAuditPIEvidenceMean}{1.0000}

\newcommand{\PIAuditPIStructuralMean}{1.0000}

\newcommand{\GenQwenClockDirectMinusBaselineWER}{-0.1292}
\newcommand{\GenQwenClockDirectMinusBaselineWERCI}{[-0.1697,-0.0895]}

\newcommand{\GenQwenNineClockDirectMinusBaselineWER}{-0.1427}
\newcommand{\GenQwenNineClockDirectMinusBaselineWERCI}{[-0.1801,-0.1058]}

\newcommand{\GenGemmaClockDirectMinusBaselineWER}{-0.1344}
\newcommand{\GenGemmaClockDirectMinusBaselineWERCI}{[-0.1695,-0.0998]}

\newcommand{\GenQwenDirectScaleControlledWER}{-0.0538}
\newcommand{\GenQwenDirectScaleControlledWERCI}{[-0.0789,-0.0292]}

\newcommand{\GenQwenDirectScaleClockWER}{-0.0730}
\newcommand{\GenQwenDirectScaleClockWERCI}{[-0.1000,-0.0467]}

\newcommand{\GenControlledMMean}{0.4953}
\newcommand{\GenControlledSwitchMean}{12.4542}
\newcommand{\GenControlledRunLengthMean}{2.6454}
\newcommand{\GenControlledRunLengthVarianceMean}{0.4096}

\newcommand{\GenControlledAmbiguityRate}{0.1208}
\newcommand{\GenClockMMean}{0.5651}
\newcommand{\GenClockSwitchMean}{14.4542}
\newcommand{\GenClockRunLengthMean}{1.7185}
\newcommand{\GenClockRunLengthVarianceMean}{1.0254}

\newcommand{\GenClockAmbiguityRate}{0.1583}

\newcommand{\PIAuditDirectSequenceEvals}{60,000}
\newcommand{\PIAuditPISequenceEvals}{602,608}
\newcommand{\PIAuditSequenceEvalRatio}{10.04}

\newcommand{\PIAuditWallTimeRatio}{4.08}

\newcommand{\PIAuditTrainableParameters}{64,929,792}

\newcommand{\HostedVFourOneMnliWER}{0.5225}
\newcommand{\HostedVFourOneMnliEvidence}{0.8091}
\newcommand{\HostedVFourOneAmiWER}{0.6910}
\newcommand{\HostedVFourOneAmiEvidence}{0.6850}
\newcommand{\HostedVFourOneIcsiWER}{0.6759}
\newcommand{\HostedVFourOneIcsiEvidence}{0.6575}
\newcommand{\HostedVFourOneReadingBankWER}{0.6339}
\newcommand{\HostedVFourOneReadingBankEvidence}{0.8127}
\newcommand{\HostedVFourOneDigitalWER}{0.8068}
\newcommand{\HostedVFourOneDigitalEvidence}{0.5167}
\newcommand{\HostedVFourOneMnliParse}{0.9284}
\newcommand{\HostedVFourOneMnliLimit}{0.0000}
\newcommand{\HostedVFourOneMnliRefusal}{0.0000}

\newcommand{\HostedVFourProMnliParse}{0.8811}
\newcommand{\HostedVFourProMnliLimit}{0.0000}
\newcommand{\HostedVFourProMnliRefusal}{0.0000}
\newcommand{\HostedQwenThreeEightMnliWER}{0.5420}
\newcommand{\HostedQwenThreeEightMnliEvidence}{0.4845}
\newcommand{\HostedQwenThreeEightAmiWER}{0.6546}
\newcommand{\HostedQwenThreeEightAmiEvidence}{0.6010}
\newcommand{\HostedQwenThreeEightIcsiWER}{0.6486}
\newcommand{\HostedQwenThreeEightIcsiEvidence}{0.5825}
\newcommand{\HostedQwenThreeEightReadingBankWER}{0.5984}
\newcommand{\HostedQwenThreeEightReadingBankEvidence}{0.6531}
\newcommand{\HostedQwenThreeEightDigitalWER}{0.7674}
\newcommand{\HostedQwenThreeEightDigitalEvidence}{0.2700}
\newcommand{\HostedQwenThreeEightMnliParse}{0.8023}
\newcommand{\HostedQwenThreeEightMnliLimit}{0.0000}
\newcommand{\HostedQwenThreeEightMnliRefusal}{0.0004}
\newcommand{\HostedQwenThreeEightMaxMnliWER}{0.3836}
\newcommand{\HostedQwenThreeEightMaxMnliEvidence}{0.6803}
\newcommand{\HostedQwenThreeEightMaxAmiWER}{0.5975}
\newcommand{\HostedQwenThreeEightMaxAmiEvidence}{0.2330}
\newcommand{\HostedQwenThreeEightMaxIcsiWER}{0.6056}
\newcommand{\HostedQwenThreeEightMaxIcsiEvidence}{0.2525}
\newcommand{\HostedQwenThreeEightMaxReadingBankWER}{0.5428}
\newcommand{\HostedQwenThreeEightMaxReadingBankEvidence}{0.6759}
\newcommand{\HostedQwenThreeEightMaxDigitalWER}{0.6171}
\newcommand{\HostedQwenThreeEightMaxDigitalEvidence}{0.2483}
\newcommand{\HostedQwenThreeEightMaxMnliParse}{0.9826}
\newcommand{\HostedQwenThreeEightMaxMnliLimit}{0.0000}
\newcommand{\HostedQwenThreeEightMaxMnliRefusal}{0.0000}

\newcommand{\HostedQwenVFourFlashMnliWER}{0.8477}
\newcommand{\HostedQwenVFourFlashMnliEvidence}{0.3742}
\newcommand{\HostedQwenVFourFlashAmiWER}{0.9937}
\newcommand{\HostedQwenVFourFlashAmiEvidence}{0.3450}
\newcommand{\HostedQwenVFourFlashIcsiWER}{1.0595}
\newcommand{\HostedQwenVFourFlashIcsiEvidence}{0.3725}
\newcommand{\HostedQwenVFourFlashReadingBankWER}{0.9530}
\newcommand{\HostedQwenVFourFlashReadingBankEvidence}{0.6074}
\newcommand{\HostedQwenVFourFlashDigitalWER}{1.1583}
\newcommand{\HostedQwenVFourFlashDigitalEvidence}{0.3300}
\newcommand{\HostedQwenVFourProMnliWER}{0.6828}
\newcommand{\HostedQwenVFourProMnliEvidence}{0.8000}
\newcommand{\HostedQwenVFourProAmiWER}{0.7598}
\newcommand{\HostedQwenVFourProAmiEvidence}{0.6040}
\newcommand{\HostedQwenVFourProIcsiWER}{0.7439}
\newcommand{\HostedQwenVFourProIcsiEvidence}{0.6175}
\newcommand{\HostedQwenVFourProReadingBankWER}{0.6535}
\newcommand{\HostedQwenVFourProReadingBankEvidence}{0.7863}
\newcommand{\HostedQwenVFourProDigitalWER}{0.8354}
\newcommand{\HostedQwenVFourProDigitalEvidence}{0.5583}

\newcommand{\StrictFreeWERMean}{0.6747}
\newcommand{\StrictFreeWERSD}{0.0242}

\newcommand{\StrictFreeEvidenceMean}{0.0814}

\newcommand{\StrictFreeParseSuccess}{0.6852}
\newcommand{\StrictFreeParseSuccessSD}{0.0274}
\newcommand{\StrictFreeLimitHit}{0.0034}
\newcommand{\StrictFreeLimitHitSD}{0.0023}
\newcommand{\StrictFreeConditionalEvidence}{0.1189}
\newcommand{\StrictFreeConditionalEvidenceSD}{0.0029}
\newcommand{\StrictFreeConditionalWER}{0.5256}
\newcommand{\StrictFreeConditionalWERSD}{0.0225}

\newcommand{\EPORMinusStrictFreeEvidenceDelta}{+0.9186}
\newcommand{\EPORMinusStrictFreeEvidenceCI}{[0.8904, 0.9444]}

\newcommand{\HistoryEporWER}{0.3757}

\newcommand{\HistoryEporEvidence}{1.0000}

\newcommand{\HistoryEporCompleteStructure}{1.0000}

\newcommand{\HistoryEporTokenAccuracy}{0.7457}

\newcommand{\HistoryEporSwitchFOne}{0.8514}

\newcommand{\HistoryAdaptedUnmaskedWER}{0.3819}

\newcommand{\HistoryAdaptedUnmaskedEvidence}{1.0000}

\newcommand{\HistoryAdaptedUnmaskedCompleteStructure}{0.9721}

\newcommand{\HistoryAdaptedUnmaskedTokenAccuracy}{0.7429}

\newcommand{\HistoryAdaptedUnmaskedSwitchFOne}{0.8492}

\newcommand{\HistoryFrozenConstrainedWER}{0.6842}
\newcommand{\HistoryFrozenConstrainedEvidence}{1.0000}
\newcommand{\HistoryFrozenConstrainedCompleteStructure}{1.0000}
\newcommand{\HistoryFrozenConstrainedTokenAccuracy}{0.5278}
\newcommand{\HistoryFrozenConstrainedSwitchFOne}{0.5333}

\newcommand{\HistoryFrozenUnmaskedWER}{1.2008}
\newcommand{\HistoryFrozenUnmaskedEvidence}{1.0000}
\newcommand{\HistoryFrozenUnmaskedCompleteStructure}{0.1473}
\newcommand{\HistoryFrozenUnmaskedTokenAccuracy}{0.3830}
\newcommand{\HistoryFrozenUnmaskedSwitchFOne}{0.0894}

\newcommand{\HistoryFixedTemplateInferenceWER}{0.9126}

\newcommand{\HistoryFixedTemplateInferenceEvidence}{1.0000}

\newcommand{\HistoryFixedTemplateInferenceCompleteStructure}{1.0000}

\newcommand{\HistoryFixedTemplateInferenceTokenAccuracy}{0.4722}

\newcommand{\HistoryFixedTemplateInferenceSwitchFOne}{0.5057}

\newcommand{\HistoryCorruptTwentyFiveWER}{0.4761}
\newcommand{\HistoryCorruptTwentyFiveEvidence}{1.0000}
\newcommand{\HistoryCorruptTwentyFiveCompleteStructure}{1.0000}
\newcommand{\HistoryCorruptTwentyFiveTokenAccuracy}{0.6702}
\newcommand{\HistoryCorruptTwentyFiveSwitchFOne}{0.7763}

\newcommand{\HistoryCorruptFiftyWER}{0.5335}
\newcommand{\HistoryCorruptFiftyEvidence}{1.0000}
\newcommand{\HistoryCorruptFiftyCompleteStructure}{1.0000}
\newcommand{\HistoryCorruptFiftyTokenAccuracy}{0.6263}
\newcommand{\HistoryCorruptFiftySwitchFOne}{0.7080}

\newcommand{\HistoryCorruptHundredWER}{0.5941}
\newcommand{\HistoryCorruptHundredEvidence}{1.0000}
\newcommand{\HistoryCorruptHundredCompleteStructure}{1.0000}
\newcommand{\HistoryCorruptHundredTokenAccuracy}{0.5796}
\newcommand{\HistoryCorruptHundredSwitchFOne}{0.5692}
\newcommand{\HistoryEporReferenceWER}{0.3729}

\newcommand{\MatchedFixedTemplateWER}{0.4486}

\newcommand{\MatchedFixedTemplateMinusEporWERDelta}{+0.0730}
\newcommand{\MatchedFixedTemplateMinusEporWERCI}{[0.0620, 0.0840]}
\newcommand{\MatchedFixedTemplateEvidence}{1.0000}

\newcommand{\MatchedFixedTemplateCompleteStructure}{1.0000}

\newcommand{\MatchedFixedTemplateTokenAccuracy}{0.7113}

\newcommand{\MatchedFixedTemplateMinusEporTokenAccuracyDelta}{-0.0345}
\newcommand{\MatchedFixedTemplateMinusEporTokenAccuracyCI}{[-0.0419, -0.0271]}
\newcommand{\MatchedFixedTemplateSwitchFOne}{0.7603}

\newcommand{\MatchedFixedTemplateMinusEporSwitchFOneDelta}{-0.0911}
\newcommand{\MatchedFixedTemplateMinusEporSwitchFOneCI}{[-0.0982, -0.0842]}

\newcommand{\SimilarityLowWER}{0.3419}
\newcommand{\SimilarityLowWERCi}{[0.3227, 0.3619]}

\newcommand{\SimilarityLowTokenAcc}{0.7747}
\newcommand{\SimilarityLowTokenAccCi}{[0.7610, 0.7882]}
\newcommand{\SimilarityLowSwitchFOne}{0.8657}
\newcommand{\SimilarityLowSwitchFOneCi}{[0.8515, 0.8790]}
\newcommand{\SimilarityMediumWER}{0.3775}
\newcommand{\SimilarityMediumWERCi}{[0.3513, 0.4035]}

\newcommand{\SimilarityMediumTokenAcc}{0.7494}
\newcommand{\SimilarityMediumTokenAccCi}{[0.7317, 0.7673]}
\newcommand{\SimilarityMediumSwitchFOne}{0.8451}
\newcommand{\SimilarityMediumSwitchFOneCi}{[0.8259, 0.8641]}
\newcommand{\SimilarityHighWER}{0.3888}
\newcommand{\SimilarityHighWERCi}{[0.3665, 0.4115]}

\newcommand{\SimilarityHighTokenAcc}{0.7323}
\newcommand{\SimilarityHighTokenAccCi}{[0.7151, 0.7491]}
\newcommand{\SimilarityHighSwitchFOne}{0.8487}
\newcommand{\SimilarityHighSwitchFOneCi}{[0.8358, 0.8613]}
\newcommand{\SimilarityExtremeWER}{0.3944}
\newcommand{\SimilarityExtremeWERCi}{[0.3730, 0.4163]}

\newcommand{\SimilarityExtremeTokenAcc}{0.7266}
\newcommand{\SimilarityExtremeTokenAccCi}{[0.7100, 0.7428]}
\newcommand{\SimilarityExtremeSwitchFOne}{0.8461}
\newcommand{\SimilarityExtremeSwitchFOneCi}{[0.8298, 0.8616]}
\newcommand{\SimilarityExtremeMinusLowWER}{+0.0524}
\newcommand{\SimilarityExtremeMinusLowWERCi}{[0.0230, 0.0818]}

\newcommand{\SimilarityExtremeMinusLowTokenAcc}{-0.0481}
\newcommand{\SimilarityExtremeMinusLowTokenAccCi}{[-0.0693, -0.0268]}
\newcommand{\SimilarityExtremeMinusLowSwitchFOne}{-0.0196}
\newcommand{\SimilarityExtremeMinusLowSwitchFOneCi}{[-0.0406, 0.0013]}

\newcommand{\TurnMarkerWERMean}{0.4271}

\newcommand{\TurnMarkerParseSuccessMean}{0.9831}
\newcommand{\TurnMarkerParseSuccessSD}{0.0037}
\newcommand{\TurnMarkerLimitHitMean}{0.0010}
\newcommand{\TurnMarkerLimitHitSD}{0.0011}

\title{Word-Level Text Unmixing via Evidence-Preserving Ownership Routing with Language Models}

\author{Jinglin He}
\author{Siyang Jiang}
\author{Lixing He}
\author{Guoliang Xing}
\author{Hongkai Chen\thanks{Corresponding author.}}
\affil{The Chinese University of Hong Kong}

\iclrfinalcopy
\begin{document}
\maketitle
\addtocontents{toc}{\protect\setcounter{tocdepth}{-1}}

\begin{abstract}
Text from multiple sources can become interleaved into a single sequence when attribution metadata is lost, such as overlapping speech transcripts, complex document reading flows, or concurrent agent streams. 
While all observed words remain present, their underlying source ownership is unknown.
We formalize this challenge as \textbf{Word-Level Text Unmixing}: given an interleaved lexical stream and the source count $K$, recover the original source sequences while preserving every word occurrence and its within-source order exactly.
Conventional large language model (LLM) paradigms approach this by generating separated source texts directly.
However, coupling latent ownership inference with lexical regeneration frequently causes LLMs to omit, duplicate, or hallucinate words, violating the exact-reconstruction objective.
To resolve this, we propose \textbf{Evidence-Preserving Ownership Routing (\method)}, a framework that clearly separates ownership prediction from sequence reconstruction.
During training, \method adapts a causal LLM to predict canonical ownership routes conditioned on the interleaved stream and prior routing decisions, inducing dynamic, history-conditioned source contexts without lexical regeneration.
At inference, \method pairs completion-safe constrained decoding with deterministic indexed reconstruction, guaranteeing structurally valid $K$-source partitions that preserve every observed token occurrence exactly once.
For systematic evaluation, we introduce \textbf{\benchmark}, spanning controlled synthetic mixtures, timestamp-derived speech from AMI and ICSI, layout-derived document streams from ReadingBank, and simulated concurrent digital outputs.
Across five evaluation tracks, a 4B \method model achieves the lowest minimum-permutation word error rate (MP-WER) among finetuned baselines, reducing the five-track mean by 22.3\% relative to compact source-array generation and competing favorably with zero-shot frontier LLMs.
These findings indicate that when lexical evidence is fully observed, decoupling ownership routing from sequence reconstruction can provide a more reliable and structurally sound alternative to generative decoders.
\end{abstract}

\section{Introduction}
\label{sec:introduction}
% Text from different speakers, document reading flows, or text generators may become interleaved when separate streams are serialized into a single sequence.
% \textbf{Temporal interleaving}: In speech transcription, recognized words must be associated with speakers, and attribution errors are particularly common near speaker turns and overlapping speech. 
% Unmixing the per-speaker word sequences is therefore important for readable speaker-attributed transcripts and speaker-aware analysis ~\citep{paturi23_interspeech,wang24h_interspeech}.
% \textbf{Spatial interleaving}: In document parsing, OCR or PDF extraction may return text in an incorrect reading order, which can disrupt information extraction and other document-understanding tasks. 
% Recovering underlying reading flows restores coherent source sequences~\citep{wang-etal-2021-layoutreader,zhang-etal-2023-reading}.
% \textbf{Concurrent interleaving}: Concurrent LLM systems can likewise involve agents exchanging messages or streaming partial outputs. If an aggregation or logging layer preserves arrival order but loses provenance, recovering the per-agent streams becomes necessary for auditing or downstream orchestration \citep{wu2024autogen,zhang2026sparse,wang2025staircase}.
Text from multiple sources frequently becomes interleaved into a single sequence when attribution metadata is lost during downstream processing.
Although the observed words remain intact, the mapping between each word occurrence and its original source is unknown. 
Recovering these latent source assignments is critical for restoring readable transcripts, coherent document structures, and interpretable digital records across several primary settings. 
In \textbf{temporal interleaving} for speech processing, transcribed words from multiple speakers merge into a single chronological stream due to overlaps or interruptions, making source attribution essential for speaker-aware analysis~\citep{paturi23_interspeech,wang24h_interspeech}.
% words from multiple speakers can be merged into a single chronological stream after transcription. This occurs when speakers overlap, provide short backchannel responses, or interrupt and resume each other's utterances. As a result, words from the same speaker may become separated by words from others, making source attribution essential for producing speaker-attributed transcripts and enabling downstream speaker-aware analysis~\citep{paturi23_interspeech,wang24h_interspeech}.
In \textbf{spatial interleaving}, document extraction systems often serialize two-dimensional layouts like multi-column pages into an incorrect linear order~\citep{wang-etal-2021-layoutreader,zhang-etal-2023-reading}.
% For example, multi-column documents, forms, or pages containing parallel text regions can produce streams where neighboring words originate from different reading flows. Recovering these latent reading sequences is important for reliable document understanding and information extraction~\citep{wang-etal-2021-layoutreader,zhang-etal-2023-reading}.
% (3) \textbf{Concurrent interleaving}: 
Furthermore, \textbf{concurrent interleaving} occurs in digital systems when collaborative agents or unlabelled event logs emit partial outputs into a shared stream where only the arrival order is preserved~\citep{wu2024autogen,wang2025staircase}.
% In digital systems, multiple producers may write content into a shared stream while only preserving arrival order. For example, concurrent LLM agents may emit partial outputs through a common interface, and collaborative systems or message logs may record events without retaining their original producer identities. Recovering per-source streams enables auditing, debugging, and downstream orchestration in such systems~\citep{wu2024autogen,wang2025staircase}.

\begin{figure*}[t]
\centering
\includegraphics[width=\linewidth]{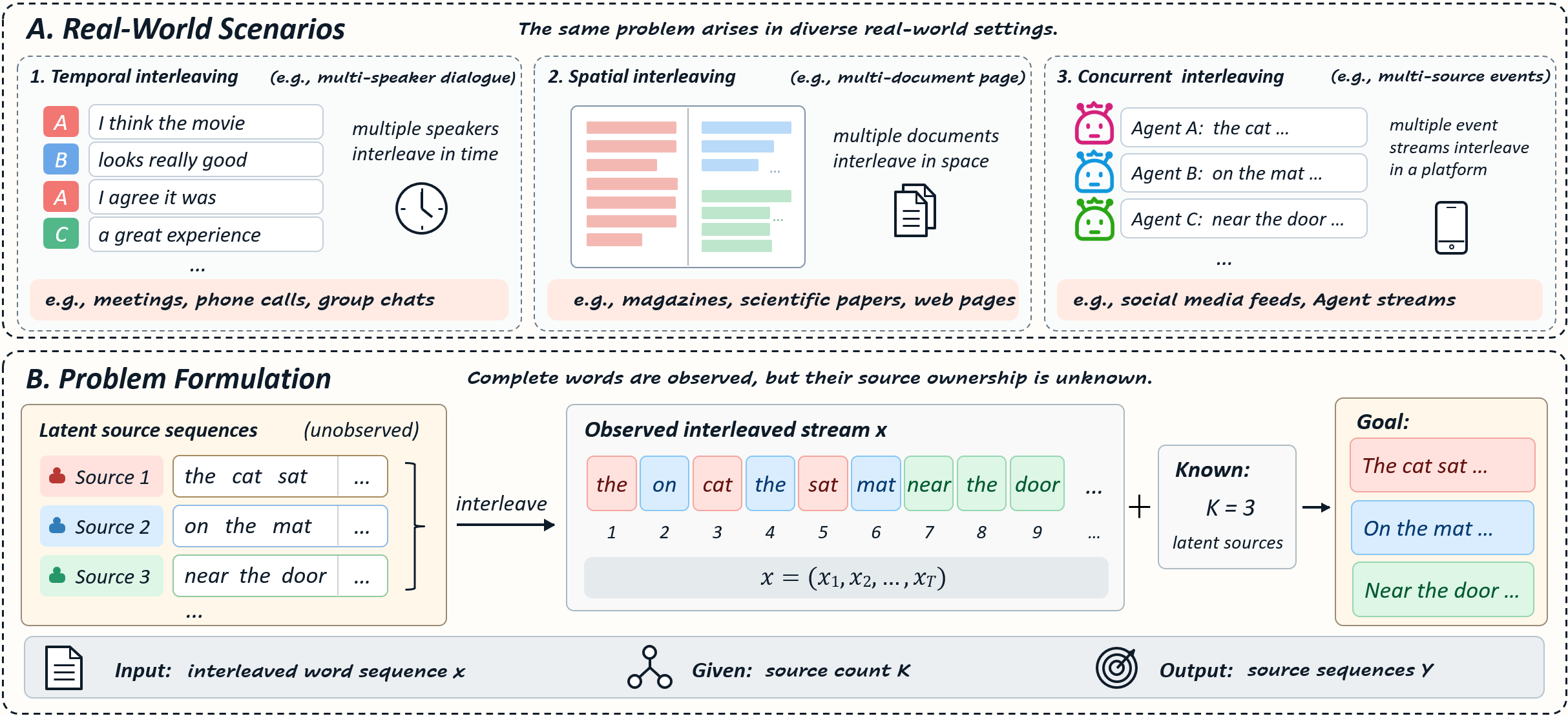}
\caption{The 
\textbf{Word-Level Text Unmixing} problem.
Temporal, spatial, and concurrent serialization motivate recovery
from an interleaved lexical stream.
Given the observed words and known source count $K$, recover the
latent sources while preserving every occurrence and its within-source order.
% \textbf{(C)} Fine-grained recovery must resolve frequent switches, similar
% source content, and exchangeable source identities in a large assignment
% space.
% The recovered source sequences are the task output; a complete source
% assignment is an equivalent latent representation used by structured systems.
}
\label{fig:text-unmixing-task}
% \vspace{-10pt}
\end{figure*}

All these scenarios share a fundamental structure where the observed words are available but their source ownership is unknown.
% the same core structure: the observed words are available, but their source ownership is not.
We formalize this problem as \textbf{Word-Level Text Unmixing}, as shown in Figure~\ref{fig:text-unmixing-task}.
Given a completely interleaved word stream and source count $K$, the objective is to recover the distinct source sequences such that every observed word occurrence is used exactly once and its original within-source order is preserved.
% recover the source sequences while using each word occurrence exactly once and preserving its order within its assigned source.
% Related tasks recover source structure at other granularities or from richer inputs. 
% Conversation disentanglement groups pre-segmented messages into conversational threads, often using speaker identities, timestamps, mentions, and reply structure
Unlike coarser text separation tasks like utterance-level conversation disentanglement~\citep{elsner-charniak-2008-talking,kummerfeld-etal-2019-large}, paragraph-level style-change detection~\citep{zangerle2022overview}, or temporally grounded speaker diarization~\citep{paturi23_interspeech,wang24h_interspeech}, word-level text unmixing requires resolving ownership shifts between adjacent words using solely the mixed lexical stream.
% Style-change detection identifies authorship changes at paragraph or sentence boundaries ~\citep{zangerle2022overview}. 
% Speaker diarization exploits acoustic and temporal cues, while language-model-based post-processing refines upstream speaker assignments ~\citep{paturi23_interspeech,wang24h_interspeech}. 
% In \problem{}, however, source ownership may change between adjacent word occurrences and only the mixed words are available.
% Text unmixing therefore requires word-level reasoning about the evolving context of each source.

While generative large language models (LLMs) are commonly used for text-to-structure extraction~\citep{lu-etal-2022-unified,wang2023instructuie,sainz-etal-2024-gollie}, directly prompting them to generate separated source arrays presents two fundamental limitations for word-level text unmixing.
% , including agent-based systems such as OneKE and KARMA~\citep{luo-etal-2025-oneke,lu-etal-2025-karma}. This motivates free-generation baselines that directly produce the separated source texts as source arrays.
% \problem{}, however, poses two fundamental challenges for this paradigm:
First, because each ownership assignment alters the partial source contexts needed to interpret subsequent words, models must perform history-conditioned routing decoupled from lexical reconstruction.
% (1) \textbf{latent ownership inference without attribution signals}. Source ownership must be inferred merely from the mixed lexical stream, without source identities or attribution metadata. Each hypothesized ownership label changes the partial source contexts used to interpret later words, requiring history-conditioned ownership prediction that is decoupled from lexical reconstruction.
Second, standard free-generation paradigms couple ownership inference with text generation. 
This allows structurally correct ownership deductions to be corrupted by the omission, duplication, or alteration of observed words, directly violating the exact-reconstruction objective.
% (2) \textbf{exact reconstruction without lexical regeneration}. Unlike conventional generation tasks, \problem{} does not require producing new text. 
% Any lexical modification introduced during generation violates the exact reconstruction objective. 
% Free generation couples ownership inference with lexical regeneration, allowing correct ownership routes to be undermined by omissions, duplications, or alterations of observed words.

To resolve this, we introduce \textbf{Evidence-Preserving Ownership Routing (\method)}, a framework that decouples learned source ownership from deterministic reconstruction.
% which separates learned source ownership from deterministic reconstruction.
During training, a causal language model predicts a single ownership label per observed word, conditioned on the interleaved stream and prior routing decisions.
% causal LLM predicts one ownership label per observed word, conditioned on the complete mixed stream and preceding labels.
This enables history-conditioned ownership prediction rather than prone-to-error lexical regeneration.
% These preceding ownership labels induce evolving partial source contexts, enabling history-conditioned ownership prediction rather than lexical regeneration.
At inference, \method a completion-safe constrained beam search to evaluate candidate routes, ensuring the selected paths can form complete \(K\)-source partitions. 
Indexed reconstruction then mechanically gathers the original word occurrences based on these routes, enforcing structural validity and exact evidence preservation.
% into source sequences in input order, ensuring structural validity by preserving every observed occurrence exactly once and maintaining within-source order. 

For systematic evaluation, we construct \textbf{\benchmark}, a benchmark covering controlled synthetic mixtures, timestamp-derived speech from AMI and ICSI, layout-derived reading flows from ReadingBank, and simulated concurrent digital streams via WikiText.
% tracks across controlled mixtures, temporal and spatial serialization derived from real metadata, and simulated digital serialization of concurrent outputs. 
% Controlled MultiNLI mixtures vary total sequence length \(T\), source count \(K\), mixing intensity \(M\), and source similarity \(S\). 
% Temporal mixtures use annotated word times from AMI and ICSI, spatial mixtures use reading-flow segments derived from ReadingBank metadata, and concurrent mixtures use virtual event-time schedules over WikiText sources. 
% Additional paired-construction and domain-transfer evaluations test generalization beyond the training construction and source corpora. 
Across the five principal evaluation tracks, a 4B \method model achieves the lowest mean minimum-permutation word error rate (\MPWER) among task-finetuned free-generation and structured baselines, yielding a 22.3\% error reduction relative to compact source-array generation.
% Its five-track mean MP-WER is 22.3\% lower than that of compact source-array generation, the stronger of the two source-array baselines.
Furthermore, \method remains highly competitive with zero-shot frontier language models.
% Moreover, the 4B \method{} system is also competitive with zero-shot frontier LLMs for \problem{}.
% Detailed ablations show that task adaptation and preceding ownership labels improve reconstruction, while constrained routing and indexed reconstruction enforce structural validity and evidence preservation independently of attribution accuracy.

Our key contributions include the following.
\begin{itemize}[leftmargin=*,nosep]
\item We formalize \textbf{\problem}, a fine-grained de-interleaving problem aimed at recovering the ownership of each word occurrence from mixed text.
% a fine-grained text de-interleaving problem that recovers the ownership of each occurrence from interleaved text.
\item We propose Evidence-Preserving Ownership Routing (\method), a structured framework that decouples history-conditioned source ownership prediction from evidence-preserving reconstruction.
% A causal LLM learns history-conditioned ownership scores from first-appearance-canonical routes, while completion-safe decoding and occurrence-indexed reconstruction enforce a valid partition and exact occurrence conservation, separating semantic source inference from evidence reconstruction.
\item We introduce \textbf{\benchmark} to evaluate text unmixing across temporal, spatial, and concurrent serialization scenarios, demonstrating \method's superior performance across multiple domains.
% a benchmark specialized for \problem{} evaluation across controlled difficulty, construction and domain shifts, and temporal, spatial, and concurrent serialization. 
% Across five principal evaluation tracks, \method{} with a 4B backbone achieves the lowest mean \MPWER{} among task-finetuned generation and structured baselines, with a 22.3\% reduction relative to compact source-array generation, competitive with zero-shot frontier LLMs.
\end{itemize}

\section{Related Work}
\label{sec:related-work}

\paragraph{Message- and utterance-level text disentanglement.}
In conversation disentanglement, most methods use lexical context,
speaker and temporal metadata, mentions, reply structure, and learned
representations
\citep{elsner-charniak-2010-disentangling,mehri2017,jiang2018,
kummerfeld-etal-2019-large,yu-joty-2020-online}.
Extensions consider zero-shot or weakly supervised objectives and explicit
speaker or reference structure
\citep{chi-rudnicky-2021-zero,zhu-etal-2021-findings,
huang-etal-2022-conversation,ma-etal-2022-structural}.
These settings retain intact linguistic units and often expose source-related
metadata \citep{chang2023,gouravajhala-etal-2023-chat}.
Style-change detection similarly locates authorship changes at paragraph or
sentence boundaries using stylistic evidence \citep{zangerle2022overview}.
Text unmixing instead reconstructs sources from individually interleaved word occurrences.

\paragraph{Lexical and multimodal speaker binding.}
Speaker separation and diarization use waveforms, timing, spatial cues, and
learned speaker representations; multimodal systems also use lexical context
\citep{chen-etal-2017-deep,horiguchi-etal-2020-eend,
park-georgiou-2018-multimodal,speakerlm-2025,
lin-etal-2025-diarization-aware}.
Lexical variants and language-model post-processing show that words can inform
fine-grained speaker assignment or correct upstream diarization labels
\citep{park-georgiou-2018-multimodal,paturi23_interspeech,
park-etal-2024-diarization-llm,wang24h_interspeech}.
These systems typically receive acoustic evidence or the initial speaker hypothesis. Instead, our task recovers the latent sources without acoustic cues or the upstream speaker hypothesis and requires exact preservation of the observed word occurrences. 
Related formulations also arise beyond NLP, including shuffle recognition and string unshuffling, document reading-order recovery, and blind source separation. We discuss these broader connections and their distinctions from \problem{} in Appendix~\ref{app:broader_connections}.

% \paragraph{Shuffle languages and string unshuffling.}
% Interleaving has also been studied in formal-language theory through the shuffle product. When the component strings are given, recognizing whether an observed string is their order-preserving shuffle is polynomial-time for any fixed number of components, while the problem becomes NP-complete when the number of components is part of the input \citep{mansfield1983computational}. Related work on shuffle squares asks whether a string can be expressed as the shuffle of two identical but unknown strings, which is NP-complete  \citep{rizzi2013recognizing,buss2014unshuffling}. These problems characterize combinatorial membership under specified component strings or equality constraints. 
% Word-level text unmixing instead leaves all source sequences latent and must infer an occurrence-level partition from the observed lexical stream, with source labels exchangeable and no non-lexical attribution signals.

\paragraph{Exchangeable structured prediction and constrained generation.}
Permutation-invariant training optimizes over equivalent label assignments
\citep{yu-etal-2017-permutation}; autoregressive clustering conditions
assignments on earlier labels \citep{pakman-etal-2020-ncp}; restricted-growth
encodings select one representative per unlabeled partition
\citep{milne-1982-restricted}; and constrained generation restricts outputs to
a prescribed structure
\citep{hokamp-liu-2017-lexically,post-vilar-2018-fast,
scholak-etal-2021-picard,geng-etal-2023-grammar}.
These methods address label symmetry, assignment dependence, and output
constraints as general prediction problems.

% Existing task formulations operate on pre-segmented units or use source-specific
% side information, while these structured-prediction methods provide general
% representations and constraints. The remaining problem is source assignment at
% the level of individual word occurrences from text alone, with guaranteed
% exact reuse of the observed evidence.

\section{Problem Formulation}
\label{sec:problem}

\paragraph{Definition.} 
In \problem, the input consists of a sequence of indexed word occurrences
$x=(x_1,\ldots,x_T)$ and a known source count $K$, with $1\leq K\leq T$.
A word occurrence is an indivisible, corpus-defined task unit whose encoding may contain multiple LLM tokens. 
Let $z^\star$ denote the latent generating ownership route. 
For any candidate route $z=(z_1,\ldots,z_T)$, the label $z_t=k$ assigns occurrence $x_t$ to source $k$. 
Defining $[K]=\{1,\ldots,K\}$, the labeled assignment space requires surjectivity so that every source is non-empty:
\begin{equation}
\mathcal P^{\mathrm{lab}}_{T,K}
=
\left\{z\in[K]^T:\{z_1,\ldots,z_T\}=[K]\right\}.
\label{eq:labeled-assignment-space}
\end{equation}
Stable filtering reconstructs
source $k$ by reading its assigned occurrences in increasing input order:
\begin{equation}
R_k(x,z)=(x_t:z_t=k,\ t=1,\ldots,T)
\label{eq:stable-filter}
\end{equation}
% with indices read in increasing order. 
Writing
$R(x,z)=(R_1(x,z),\ldots,R_K(x,z))$, the generating source tuple is
$Y^\star=R(x,z^\star)$.
Source labels are exchangeable, meaning that for any permutation in the symmetric group 
% For any permutation $\pi$ in the symmetric
% group 
$S_K$, the tuples $Y=(y_1,\ldots,y_K)$ and
$Y^\pi=(y_{\pi(1)},\ldots,y_{\pi(K)})$ represent the same answer.
The text unmixing task therefore recovers
% \begin{equation}
$(x,K)\longmapsto [Y^\star]$,
where 
$Y^\star=(y_1^\star,\ldots,y_K^\star)$,
% \label{eq:task-output}
% \end{equation}
and brackets denote equivalence under a global source permutation. The
output is an unordered collection with multiplicity rather than a set of named
sources, and its valid output space is:
\begin{equation}
\mathcal Y_{\mathrm{valid}}(x,K)
=
\left\{[R(x,z)]:z\in\mathcal P^{\mathrm{lab}}_{T,K}\right\}.
\label{eq:valid-output-space}
\end{equation}

The only input is $(x,K)$, and no source identities, timestamps,
layout coordinates, or upstream attribution hypotheses are available, while the word occurrences themselves are complete and fixed.

\paragraph{Example.} Consider two latent source sequences, ``the volcano erupts'' and ``the chef cooks.'' 
After interleaving, the model observes only the indexed sequence
$$
x=(\text{the},\;\text{the},\;\text{volcano},\;\text{erupts},\;\text{chef},\;\text{cooks})
$$
alongside the number of sources \(K=2\). 
The \problem task is to recover the two original source sequences by determining which source owns each observed word occurrence, while using every occurrence exactly once and preserving its order within that source. 
For this scenario, the latent ownership route is
$$
z=(1,2,1,1,2,2),
$$
which groups \((x_1,x_3,x_4)\) into ``the volcano erupts'' and \((x_2,x_5,x_6)\) into ``the chef cooks.''
An ownership route is a sequence of source labels for the observed occurrences. 
The model does not observe \(z\) but rather  predicts this auxiliary representation in order to reconstruct the sources.

\paragraph{Validity and correctness.}
Structural validity strictly requires exactly $K$ non-empty sources, exact evidence preservation, and unchanged word order within each assigned source.
Evidence preservation dictates that every observed word occurrence is reused exactly once.
Semantic correctness evaluates the logical and semantic validity of each resulting unmixed source sequence.
% Identifiability refers to a concern that distinct partitions can yield different, linguistically plausible source collections for the same mixed sequence.

% This formulation separates the learned and known parts of the problem.
% Ownership requires a scorer that uses the mixed stream and its evolving
% assignments, whereas occurrence conservation and source-local order follow
% from deterministic indexing. \method{} instantiates this division by scoring
% ownership routes and applying a canonical, completion-safe reconstruction
% interface to every selected route.

\section{Approach: Evidence-Preserving Ownership Routing}
\label{sec:method}

% \method{} follows three requirements of the task in Section~\ref{sec:problem}.
% Exchangeable source labels call for a canonical representation.
% Ownership decisions depend on earlier assignments, motivating a
% history-conditioned scorer. Exact reuse of the observed word occurrences
% requires constrained ownership routes and indexed reconstruction.
% A causal LM therefore predicts the latent route $z$, from which the source
% sequences $Y$ are reconstructed. Figure~\ref{fig:epor-overview} summarizes
% training and inference. For any accepted route, evidence preservation and
% structural validity hold independently of ownership accuracy.

\begin{figure}[t]
  \centering
  \includegraphics[width=\linewidth]{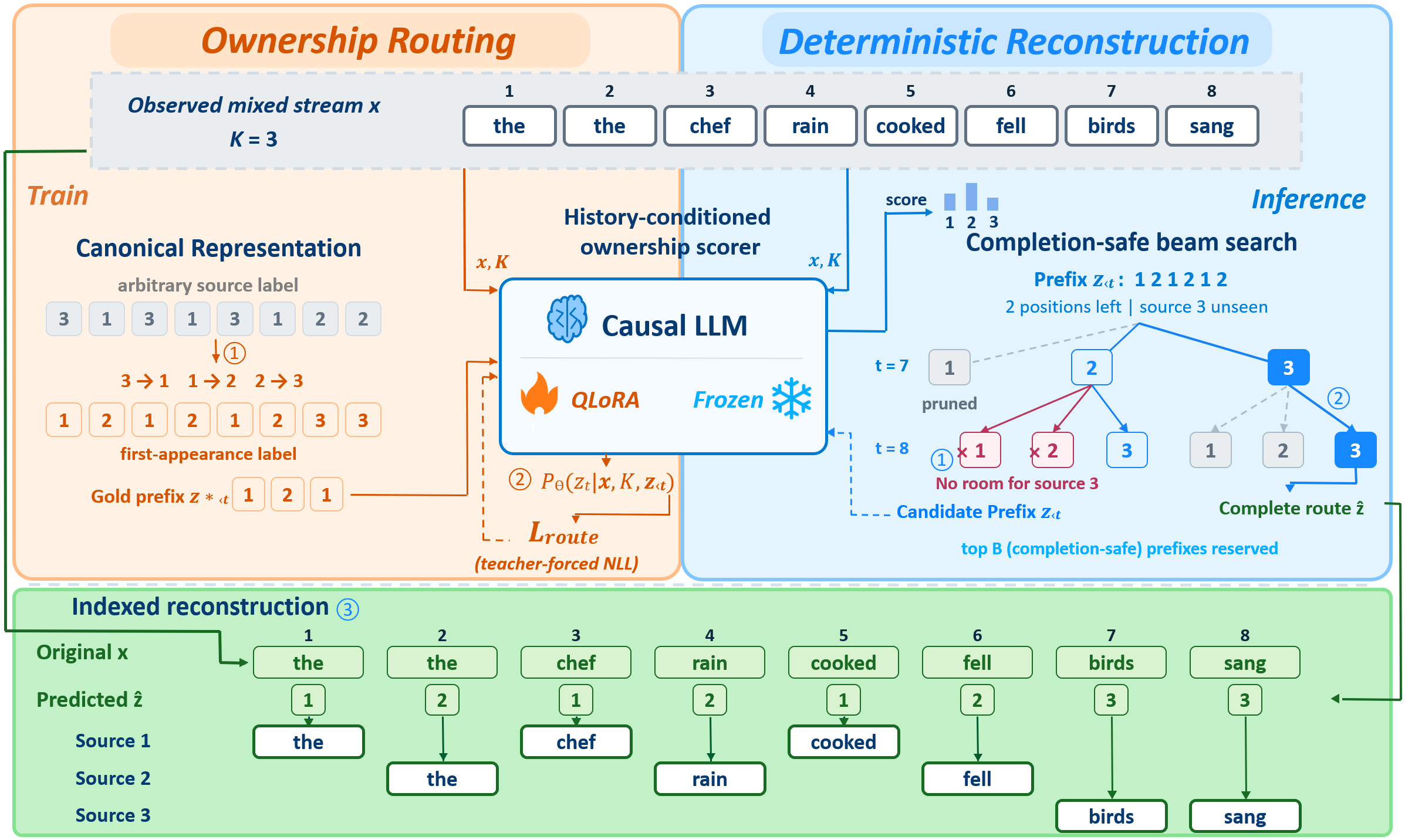}
  \caption{\textbf{Evidence-Preserving Ownership Routing (\method{}).}
  \textbf{Training:} the causal LLM learns canonical ownership from the complete
  mixed input and gold prefixes.
  \textbf{Inference:} completion-safe beam search scores candidate
  ownership routes, and then indexed reconstruction gathers the original occurrences
  into sources.}
  \label{fig:epor-overview}
\end{figure}

\method{} separates learned source ownership from evidence preservation. During training, a causal LLM predicts an ownership label for each observed word, conditioned on the complete mixed stream, the source count, and preceding labels. These labels define partial source contexts that guide subsequent predictions.
At inference, the model scores candidate ownership routes, while completion-safe beam search retains prefixes that can form a partition into \(K\) non-empty sources. Indexed reconstruction then selects the original occurrences in source-local order. Figure~\ref{fig:epor-overview} shows both stages.

\subsection{Training: Ownership-Only Adaptation}
\label{subsec:epor-training}

For each example $(x,K,z^\star)$, we canonicalize the gold route by source
order of first appearance and adapt the causal LLM with QLoRA under teacher-forced negative log-likelihood. 
$z^\star$ denotes the canonical gold route.
Algorithm~\ref{alg:epor-training} in Appendix~\ref{app:algorithms} gives the
procedure.

\subsubsection{Canonical Representation}
\label{subsec:canonical-quotient}

Because source labels are exchangeable, we use a first-appearance canonical ownership representation.
The source containing the earliest occurrence is labeled 1, the next unseen source is labeled 2, and so on. 
This restricted-growth representation
\citep{milne-1982-restricted} defines
\begin{equation}
\mathcal Z_{T,K}
=
\left\{
\begin{array}{@{}l@{}}
z\in\{1,\ldots,K\}^{T}:\ z_1=1,\\
z_t\leq 1+\max_{j<t}z_j\ \text{for }t>1,\quad \max_t z_t=K
\end{array}
\right\}.
\label{eq:legal-rgs}
\end{equation}
Each unlabeled partition into $K$ non-empty sources has one such route,
replacing $K!$ equivalent labelings with a shared target for supervision and
decoding. This removes naming redundancy without resolving semantic
ambiguity (Appendix~\ref{app:support-proofs}).

\subsubsection{History-Conditioned Ownership Adaptation}
\label{subsec:history-score}

The learning target is the canonical ownership route $z^\star$. Given the complete indexed stream $x$ and source count $K$, the LLM predicts one ownership label per word occurrence using its existing vocabulary head.

The training scorer factorizes autoregressively:
\begin{equation}
s_\theta(z;x,K)
=
\sum_{t=1}^{T}\log p_\theta(z_t\mid x,K,z_{<t}).
\label{eq:epor-score}
\end{equation}
Here $p_\theta(z_t\mid\cdot)$ is the full-vocabulary LLM probability of the symbol encoding $z_t$. 
During training, each label prediction is conditioned on the gold route prefix $z^\star_{<t}$.
Together with $x$, a prefix determines the partial sequence for each source,
\begin{equation}
H_{k,t}(x,z_{<t})=(x_j:j<t,\ z_j=k).
\end{equation}
Different prefixes associate the same occurrence with different source
contexts. The sequences $H_{k,t}$ are implicit in $(x,z_{<t})$, rather than
separate model inputs. As in autoregressive clustering
\citep{pakman-etal-2020-ncp}, the scorer can use the contexts implied by
earlier assignments.

Let $P_i$ be the prompt containing $(x_i,K_i)$, $v_i^\star$ the serialized
canonical route with its assistant termination suffix, and $\mathcal U_i$
the supervised LM-token positions. The QLoRA adaptation objective is
\begin{equation}
\mathcal L_{\mathrm{route}}(\theta)
=
-\frac{1}{\sum_i|\mathcal U_i|}
\sum_i\sum_{u\in\mathcal U_i}
\log p_\theta(v^\star_{i,u}\mid P_i,v^\star_{i,<u}).
\label{eq:route-objective}
\end{equation}
The prompt is loss-masked, and the route and assistant termination suffix are
supervised (Appendix~\ref{app:support-proofs}). 
Because the complete $x$ precedes the route, every decision can attend to the full mixed stream.

\subsection{Inference: Constrained Ownership Routing}
\label{subsec:epor-inference}
% inference supplies each candidate's predicted prefix.

At inference, the adapted causal LLM scores each candidate ownership route using its predicted prefix. Completion-safe constrained decoding searches routes that can form $K$ non-empty sources, and then indexed reconstruction then selects the original word occurrences according to the chosen route. 
Algorithm~\ref{alg:epor-inference} in Appendix~\ref{app:algorithms} gives the complete procedure.

\subsubsection{Completion-Safe Decoding}
\label{subsec:completion-safe-decoding}

First-appearance validity alone does not ensure that all $K$ sources can
appear before the route ends. Before position $t$, let
$m_{t-1}=\max z_{<t}$, with $m_0=0$. The admissible labels are
\begin{equation}
\mathcal A_t(z_{<t})
=
\left\{
a\in\{1,\ldots,\min(K,m_{t-1}+1)\}:
K-\max(m_{t-1},a)\leq T-t
\right\}.
\label{eq:legal-action-set}
\end{equation}
The range admits an existing source or the next unseen source, so that the
remaining-capacity condition reserves enough positions for all missing
sources. If two positions and two unseen sources remain, the next label
must introduce a source. Appendix~\ref{app:completion-safe-proof} proves
that these conditions characterize exactly the extensions admitting a
complete canonical route.

\subsubsection{Finite-Beam Ownership Search}
\label{subsec:finite-beam-search}

Each prefix is expanded over $\mathcal A_t(z_{<t})$, retaining the top
$B$ extensions under Equation~\ref{eq:epor-score}. As in constrained decoding
\citep{hokamp-liu-2017-lexically,post-vilar-2018-fast,
scholak-etal-2021-picard,geng-etal-2023-grammar}, illegal actions are masked.
Scores retain full-vocabulary log probabilities without renormalization
over legal labels. Decoding stops after $T$ decisions. 

The search retains the top $B$ extensions at each step:
\begin{equation}
\widehat z_B=\operatorname{BeamSearch}_{B}(s_\theta;\mathcal Z_{T,K}).
\label{eq:epor-interface}
\end{equation}
Here $B$ is the beam width.

\subsubsection{Occurrence-Indexed Reconstruction}
\label{subsec:deterministic-reconstruction}
Stable filtering $R(x,\widehat z_B)$ in Equation~\ref{eq:stable-filter} reads
each original occurrence exactly once, in an increasing input order within its
assigned source. 
Together with completion-safe decoding, it yields $K$
non-empty source sequences after $T$ route decisions:
\begin{equation}
    \qquad \widehat Y=R(x,\widehat z_B),
\end{equation}
where $R$ is the stable-filtering map.
Indexed reconstruction preserves the observed occurrences regardless of whether the predicted ownership route is correct.
Appendix~\ref{app:reconstruction-guarantee} gives the formal guarantee.

\section{\benchmark{}}
\label{sec:benchmark}

\benchmark{} is a benchmark introduced to evaluate \problem{} across five text-interleaving tracks. Each example contains a mixed stream $x$, source count $K$,
reference sources $Y^\star$, and gold ownership route $z^\star$.
Models receive only $x$ and $K$, while source identities and construction metadata are withheld.
Task adaptation and checkpoint selection use 10,000 training and 1,000 development mixtures
from disjoint SNLI sources.

\paragraph{Five evaluation tracks.}
Table~\ref{tab:benchmark-tracks} summarizes the tracks used in the main
comparison. Controlled MultiNLI varies source count $K$, total length $T$,
mixing intensity $M$, and source similarity $S$ on held-out source groups.
Here, \emph{controlled} describes the mixture construction, and evaluation still
involves a corpus shift from SNLI. 
Temporal tracks merge AMI and ICSI words by annotated times. The spatial track derives reading-flow segments from ReadingBank XML metadata \citep{wang-etal-2021-layoutreader}. Concurrent digital serialization applies smooth, bursty, and unequal-rate virtual event
schedules to the same 400 source groups and averages the three profiles.
Appendix~\ref{app:data-construction} illustrates detailed corpus preprocessing, splits, and construction parameters.

\begin{table}[t]
\centering
\scriptsize
\setlength{\tabcolsep}{4pt}
\caption{\textbf{The \benchmark{} evaluation tracks.}
Counts give examples / provenance groups. $K$ is the source count and $T$ the
number of observed word occurrences. }
\label{tab:benchmark-tracks}
\begin{tabular*}{\textwidth}{@{\extracolsep{\fill}}llrcll@{}}
\toprule
Track & Corpus & Examples / groups & $K$ & $T$ & Interleaving \\
\midrule
Controlled & MultiNLI & 2,640 / 240 & 2--4 & 16/24/32 & Controlled intensity \\
Temporal & AMI & 1,000 / 59 & 2/3 & 64 & Word timestamps \\
Temporal & ICSI dev & 400 / 15 & 2/3 & 64 & Word timestamps \\
Spatial & ReadingBank & 833 / 828 & 2--4 & 32/64 & Reading-flow metadata \\
Concurrent & WikiText & 1,200 / 400 & 2--4 & 32/64 & Virtual event times \\
\bottomrule
\end{tabular*}
\end{table}

\paragraph{Additional transfer evaluations.}
A paired construction set holds the source text fixed and compares the
controlled mixer with an independent clock mixer on 240 further MultiNLI
groups. Domain-transfer sets use WikiText, AG News, and speech-derived text
without target-domain adaptation. AMI reuses the temporal track's
1,000-window view, and ICSI uses a separate 300-window reserve with no meetings
shared with its 400-window development track. These evaluations, unitization
rules, and the identifiability audit are specified in
Appendix~\ref{app:data-construction}.

\section{Experiments}
\label{sec:experiments}
\method{} is systematically evaluated on \benchmark{}.
We compare it with free-generation and structured baselines across five principal tracks, then test the roles of canonical representation, ownership-route history, and decoding constraints. We also examine generalization across constructions, domains, and backbones, followed by an analysis of text-unmixing difficulty. Appendix~\ref{app:more-experiments} reports additional diagnostics and transfer results.

\subsection{Experimental Setup}

\paragraph{Comparisons and baselines.}
Task-adapted systems share the Qwen3.5-4B backbone and SNLI data
\citep{qwen35-4b-model-card}. The free-generation baselines produce source-word arrays in two variants: compact source-array generation and evidence-explicit source-array generation. A separate turn-marker generator reproduces the mixed stream with source turn markers. Structured baselines---Global Transformer, EMA
prototype, RGS-CRF, and legal projection---predict or recover indexed
ownership. Frontier zero-shot models use evidence-explicit source generation.
Ablations remove task adaptation, ownership-route history, or legality
masking.
PI-Route tests permutation-invariant supervision in place of canonical labels. Appendix~\ref{app:experimental-details} details these comparisons.

\paragraph{Training and inference.}
Task adaptation uses rank-32 NF4 QLoRA for two epochs and three runs
(seeds 42--44) \citep{dettmers-etal-2023-qlora}. \method{} uses teacher-forced
route supervision and completion-safe beam search with $B=4$. No target-track
examples are used for training or checkpoint selection. Frontier calls use
non-thinking inference at temperature zero
(Appendix~\ref{app:hosted-controls}). Outputs receive no post-hoc repair
except in the designated legal-projection baseline.

\paragraph{Metrics.}
We use minimum-permutation WER (\MPWER{}), permutation-aligned ownership
accuracy and switch F1 for indexed routes, and exact word-multiset
preservation (Evidence). Structural validity follows
Section~\ref{sec:problem}, and route validity additionally requires canonical
labels. Results are macro-averaged, with paired uncertainty grouped by
provenance (Appendix~\ref{app:additional-metrics}).

\subsection{Main Comparison: Ownership Prediction versus Evidence Regeneration}
\label{sec:results}
\label{sec:reconstruction}
\begin{table*}[t]
\centering
\scriptsize
\setlength{\tabcolsep}{2.7pt}
\renewcommand{\arraystretch}{1.02}
\caption{\textbf{Main results across five evaluation tracks on \benchmark{}.} Each block
reports \MPWER{} (MP), permutation-aligned ownership accuracy (Own.), and
evidence preservation (Ev.). Turn-marker ownership uses lexical alignment
(Appendix~\ref{app:additional-metrics}).
Lower \MPWER{} is better; higher values are better for the other metrics.
Best and second-best results are bold and underlined. 
Turn-marker uncertainty is given in Table~\ref{tab:turn-marker-transfer}, and Appendix~\ref{app:hosted-controls} reports hosted-model settings.}
\label{tab:confirmation}
\resizebox{\textwidth}{!}{%
\begin{tabular}{@{}l
rrr
rrr
rrr
rrr
rrr@{}}
\toprule
&
\multicolumn{3}{c}{\textbf{MNLI}} &
\multicolumn{3}{c}{\textbf{AMI}} &
\multicolumn{3}{c}{\textbf{ICSI-dev}} &
\multicolumn{3}{c}{\textbf{RB}} &
\multicolumn{3}{c}{\textbf{DIG}} \\
\cmidrule(lr){2-4}
\cmidrule(lr){5-7}
\cmidrule(lr){8-10}
\cmidrule(lr){11-13}
\cmidrule(lr){14-16}

\textbf{System}
& \textbf{MP}$\downarrow$ & \textbf{Own.}$\uparrow$ & \textbf{Ev.}$\uparrow$
& \textbf{MP}$\downarrow$ & \textbf{Own.}$\uparrow$ & \textbf{Ev.}$\uparrow$
& \textbf{MP}$\downarrow$ & \textbf{Own.}$\uparrow$ & \textbf{Ev.}$\uparrow$
& \textbf{MP}$\downarrow$ & \textbf{Own.}$\uparrow$ & \textbf{Ev.}$\uparrow$
& \textbf{MP}$\downarrow$ & \textbf{Own.}$\uparrow$ & \textbf{Ev.}$\uparrow$ \\
\midrule

\multicolumn{16}{@{}l}{\itshape Generative baselines} \\
\addlinespace[1pt]

Compact source-array generation (4B)
& \ConfirmFreeWERMean & -- & \ConfirmFreeEvidenceMean
& .7275 & -- & .0423
& .7439 & -- & .0400
& .8415 & -- & .0240
& .7310 & -- & .0464 \\

Evidence-explicit source-array generation (4B)
& \StrictFreeWERMean & -- & \StrictFreeEvidenceMean
& .7418 & -- & .0343
& .7530 & -- & .0317
& .8547 & -- & .0272
& .7314 & -- & .0544 \\

\ifturnmarkerresults
Turn-marker generation (4B)
& 0.4271 & \underline{0.6991} & 0.3957
& 0.6928 & 0.4441 & 0.6377
& 0.6690 & 0.4805 & \underline{0.6600}
& 0.6862 & 0.4959 & 0.3325
& \underline{0.6449} & 0.5426 & \underline{0.7769} \\
\fi

\midrule

\multicolumn{16}{@{}l}{\itshape Structured baselines} \\
\addlinespace[1pt]

Global Transformer
& \ConfirmGlobalWERMean
& \ConfirmGlobalTokenAccMean
& \textbf{\ConfirmGlobalEvidenceMean}
& .6406 & .6086 & \textbf{1.0000}
& .6389 & .6075 & \textbf{1.0000}
& .6820 & .5866 & \textbf{1.0000}
& .7321 & .5397 & \textbf{1.0000} \\

EMA prototype
& \ConfirmEmaWERMean
& \ConfirmEmaTokenAccMean
& \textbf{\ConfirmEmaEvidenceMean}
& .6596 & .6201 & \textbf{1.0000}
& .6674 & .6136 & \textbf{1.0000}
& .7427 & .5814 & \textbf{1.0000}
& .7078 & .5482 & \textbf{1.0000} \\

RGS-CRF
& \ConfirmCrfWERMean
& \ConfirmCrfTokenAccMean
& \textbf{\ConfirmCrfEvidenceMean}
& .6483 & .6108 & \textbf{1.0000}
& .6495 & .6076 & \textbf{1.0000}
& .6555 & .5989 & \textbf{1.0000}
& .7032 & .5552 & \textbf{1.0000} \\

Legal projection
& \ConfirmProjectionWERMean
& \ConfirmProjectionTokenAccMean
& \textbf{\ConfirmProjectionEvidenceMean}
& .6431 & \textbf{.6255} & \textbf{1.0000}
& .6483 & \underline{.6211} & \textbf{1.0000}
& .6843 & \underline{.6015} & \textbf{1.0000}
& .6764 & \underline{.5712} & \textbf{1.0000} \\

\midrule

\multicolumn{16}{@{}l}{\itshape Zero-shot frontier LLMs} \\
\addlinespace[1pt]

Qwen3.8-Flash (125B)
& \HostedQwenThreeEightMnliWER & -- & \HostedQwenThreeEightMnliEvidence
& \HostedQwenThreeEightAmiWER & -- & \HostedQwenThreeEightAmiEvidence
& \HostedQwenThreeEightIcsiWER & -- & \HostedQwenThreeEightIcsiEvidence
& \underline{\HostedQwenThreeEightReadingBankWER} & -- & \HostedQwenThreeEightReadingBankEvidence
& \HostedQwenThreeEightDigitalWER & -- & \HostedQwenThreeEightDigitalEvidence \\

Qwen3.8-Max (2.4T)
& \underline{\HostedQwenThreeEightMaxMnliWER} & -- & \HostedQwenThreeEightMaxMnliEvidence
& \textbf{\HostedQwenThreeEightMaxAmiWER} & -- & \HostedQwenThreeEightMaxAmiEvidence
& \textbf{\HostedQwenThreeEightMaxIcsiWER} & -- & \HostedQwenThreeEightMaxIcsiEvidence
& \textbf{\HostedQwenThreeEightMaxReadingBankWER} & -- & \HostedQwenThreeEightMaxReadingBankEvidence
& \textbf{\HostedQwenThreeEightMaxDigitalWER} & -- & \HostedQwenThreeEightMaxDigitalEvidence \\

DeepSeek-V4-Flash (284B)
& \HostedQwenVFourFlashMnliWER & -- & \HostedQwenVFourFlashMnliEvidence
& \HostedQwenVFourFlashAmiWER & -- & \HostedQwenVFourFlashAmiEvidence
& \HostedQwenVFourFlashIcsiWER & -- & \HostedQwenVFourFlashIcsiEvidence
& \HostedQwenVFourFlashReadingBankWER & -- & \HostedQwenVFourFlashReadingBankEvidence
& \HostedQwenVFourFlashDigitalWER & -- & \HostedQwenVFourFlashDigitalEvidence \\

DeepSeek-V4-Pro (1.6T)
& \HostedQwenVFourProMnliWER & -- & \HostedQwenVFourProMnliEvidence
& \HostedQwenVFourProAmiWER & -- & \HostedQwenVFourProAmiEvidence
& \HostedQwenVFourProIcsiWER & -- & \HostedQwenVFourProIcsiEvidence
& \HostedQwenVFourProReadingBankWER & -- & \HostedQwenVFourProReadingBankEvidence
& \HostedQwenVFourProDigitalWER & -- & \HostedQwenVFourProDigitalEvidence \\

DeepSeek-V4.1-Flash (552B)
& \HostedVFourOneMnliWER & -- & \underline{\HostedVFourOneMnliEvidence}
& \HostedVFourOneAmiWER & -- & \underline{\HostedVFourOneAmiEvidence}
& \HostedVFourOneIcsiWER & -- & \HostedVFourOneIcsiEvidence
& \HostedVFourOneReadingBankWER & -- & \underline{\HostedVFourOneReadingBankEvidence}
& \HostedVFourOneDigitalWER & -- & \HostedVFourOneDigitalEvidence \\

\midrule

\multicolumn{16}{@{}l}{\itshape Proposed method} \\
\addlinespace[1pt]

\textbf{\method{} (4B; ours)}
& \textbf{\ConfirmDirectRgsWERMean}
& \textbf{\ConfirmDirectRgsTokenAccMean}
& \textbf{\ConfirmDirectRgsEvidenceMean}
& \underline{.6320} & \underline{.6205} & \textbf{1.0000}
& \underline{.6161} & \textbf{.6228} & \textbf{1.0000}
& .6156 & \textbf{.6224} & \textbf{1.0000}
& .6474 & \textbf{.5912} & \textbf{1.0000} \\

\bottomrule
\end{tabular}%
}
\end{table*}

Among task-adapted systems, \method{} has the lowest mean \MPWER{} on
four of five tracks and the lowest five-track average
(see Table~\ref{tab:confirmation}). On MultiNLI, it reduces \MPWER{} from
0.6696 to 0.3757 relative to compact source-array generation, a 43.9\%
relative reduction. Evidence-explicit instructions do not close this gap;
the low Evidence scores reflect imperfect lexical reproduction
(Appendix~\ref{app:generation-diagnostics}).

\ifturnmarkerresults
Generating words in mixed-stream order narrows the gap. Turn-marker
generation reaches \TurnMarkerWERMean{} \MPWER{} on MultiNLI, but
\method{} remains better on MultiNLI, AMI, ICSI, and ReadingBank, with all
four paired 95\% intervals excluding zero. On Digital, turn-marker
generation is slightly lower (0.6449 versus 0.6474); the paired
\method{}-minus-turn-marker difference is
$+0.0026$ (95\% CI $[-0.0067, 0.0120]$). \method{} has higher ownership accuracy
on all five tracks and preserves evidence exactly, whereas turn-marker
Evidence ranges from 0.3325 to 0.7769. Mixed-order generation therefore
reduces reordering demands but still introduces lexical errors
(Appendix~\ref{app:turn-marker}).
\fi

All structured baselines preserve evidence exactly, and their errors concern source ownership.
\method{} reaches 0.7457 ownership accuracy on MultiNLI, though legal projection is more accurate on AMI. 
The 4B \method{} system is competitive with frontier zero-shot models in reconstruction, with the best \MPWER{} varying by track. Appendix~\ref{app:hosted-controls} reports
their configurations and failure diagnostics.

\subsection{Ablation and Component Analysis}
\label{sec:history-results}

\begin{table*}[t]
\centering
% \footnotesize
\scriptsize
\setlength{\tabcolsep}{2pt}
\renewcommand{\arraystretch}{1.02}
\caption{\textbf{EPOR component and representation ablations on MultiNLI.}
Adapt. denotes task adaptation; Legal mask denotes completion-safe decoding.
PI-Route is a matched representation control with noncanonical legal support,
rather than a component-removal condition.}
\label{tab:mechanism-full}
\begin{tabular*}{\textwidth}{@{}lcc@{\extracolsep{\fill}}rrrrr@{}}
\toprule
Condition & Adapt. & \shortstack{Legal mask} & \MPWER{} $\downarrow$ & Evidence $\uparrow$ & \shortstack{Complete $K$-way $\uparrow$} & \shortstack{Ownership acc. $\uparrow$} & \shortstack{Switch F1 $\uparrow$} \\
\midrule
\method{} & $\checkmark$ & $\checkmark$ & \HistoryEporWER{} & \HistoryEporEvidence{} & \HistoryEporCompleteStructure{} & \HistoryEporTokenAccuracy{} & \HistoryEporSwitchFOne{} \\
w/o task adaptation & $\times$ & $\checkmark$ & \HistoryFrozenConstrainedWER{} & \HistoryFrozenConstrainedEvidence{} & \HistoryFrozenConstrainedCompleteStructure{} & \HistoryFrozenConstrainedTokenAccuracy{} & \HistoryFrozenConstrainedSwitchFOne{} \\
w/o legality masking & $\checkmark$ & $\times$ & \HistoryAdaptedUnmaskedWER{} & \HistoryAdaptedUnmaskedEvidence{} & \HistoryAdaptedUnmaskedCompleteStructure{} & \HistoryAdaptedUnmaskedTokenAccuracy{} & \HistoryAdaptedUnmaskedSwitchFOne{} \\
\shortstack[l]{w/o adaptation and\\legality masking} & $\times$ & $\times$ & \HistoryFrozenUnmaskedWER{} & \HistoryFrozenUnmaskedEvidence{} & \HistoryFrozenUnmaskedCompleteStructure{} & \HistoryFrozenUnmaskedTokenAccuracy{} & \HistoryFrozenUnmaskedSwitchFOne{} \\
\shortstack[l]{Route-independent fixed-\\template history} & $\checkmark$ & $\checkmark$ & \MatchedFixedTemplateWER{} & \MatchedFixedTemplateEvidence{} & \MatchedFixedTemplateCompleteStructure{} & \MatchedFixedTemplateTokenAccuracy{} & \MatchedFixedTemplateSwitchFOne{} \\
\midrule
\multicolumn{8}{c}{\itshape Representation control}\\
PI-Route & $\checkmark$ & $\checkmark$ & \PIAuditPIWERMean{} & \PIAuditPIEvidenceMean{} & \PIAuditPIStructuralMean{} & \PIAuditPITokenAccMean{} & \PIAuditPISwitchFOneMean{} \\
\midrule
\multicolumn{8}{c}{\itshape Interface baseline}\\
\shortstack[l]{Evidence-explicit\\source-array generation} & $\checkmark$ & $\times$ & \StrictFreeWERMean{} & \StrictFreeEvidenceMean{} & -- & -- & -- \\
\bottomrule
\end{tabular*}
\end{table*}

\paragraph{Canonical representation.}
The matched PI-Route control replaces canonical supervision with exhaustive
permutation-invariant training. It obtains \PIAuditPIWERMean{} \MPWER{}
versus \PIAuditDirectWERMean{} for \method{}
(Table~\ref{tab:mechanism-full}), with paired recovery intervals spanning zero.
Its training requires \PIAuditSequenceEvalRatio{} times as many LLM sequence evaluations and \PIAuditWallTimeRatio{} times the wall time in the measured
implementation. 
Canonicalization thus removes permutation redundancy at substantially lower training cost, with comparable recovery quality.
Appendix~\ref{app:pi-route} gives the full comparison and cost accounting.

\paragraph{Assignment-specific history.}
Replacing ownership-route prefixes with a route-independent fixed template during
training and evaluation raises \MPWER{} from 0.3757 to 0.4486 and lowers
ownership accuracy from 0.7457 to 0.7113. Evidence and complete $K$-way
structure remain unchanged, where legal actions still depend on each candidate's
actual prefix, isolating the contribution of model-visible assignment
history. Appendix~\ref{app:mechanism-controls} also reports a graded
history-corruption analysis.

\paragraph{Legality constraints and task adaptation.}
Removing the full legality mask after adaptation changes \MPWER{} from
0.3757 to 0.3819 and reduces complete $K$-way outputs from 1.0000 to 0.9721.
The mask chiefly guarantees structure. With the same completion-safe decoder, task adaptation reduces \MPWER{} from 0.6842 to 0.3757.

\paragraph{Finite beam search.}
Increasing beam width from 1 to 32 lowers selected-route \MPWER{} from
\MechanismBeamOneSelectedWER{} to \MechanismBeamThirtyTwoSelectedWER{}.
This gold-aware diagnostic, unavailable at inference, reveals substantial
ranking error even among legal candidates
(Appendix~\ref{app:beam-search}).

\subsection{Generalization and Transfer}
\label{sec:generalization}

Ownership prediction outperforms source-array generation under an independent
clock-based construction for Qwen3.5-4B, Qwen3.5-9B, and Gemma-7B-IT.
These cross-backbone checks use one training run per backbone. 
In domain transfer, \method{} improves over RGS-CRF on WikiText,
the shared AMI view, and the separate ICSI reserve, but performs worse on
AG News. Appendix~\ref{app:construction-backbone}--\ref{app:real-serialization-results}
gives construction, domain, and per-view serialization results.

\subsection{What Makes Text Unmixing Difficult?}
\label{sec:difficulty}

\begin{figure}[t]
  \centering
  \includegraphics[width=0.72\linewidth]{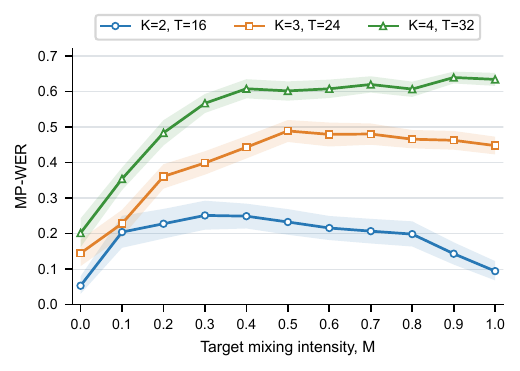}
  \caption{\textbf{Recovery across controlled mixing intensities.} The same
  source groups are evaluated at eleven target values of $M$. Points average
  three training runs and 80 source groups per $(K,T)$ condition; shaded
  regions show source-group bootstrap 95\% confidence intervals.}
  \label{fig:mixing-full-grid}
\end{figure}

Our controlled analyses point to two distinct sources of difficulty: ambiguity in the ownership route and similarity between the latent sources. Route ambiguity is not monotonic in mixing intensity, as shown in Figure~\ref{fig:mixing-full-grid}. With two sources, maximal switching forces an alternating route and can therefore make the ownership structure easier to recover, whereas intermediate mixing permits a less constrained set of assignments. Recovery is also worse for source groups with higher frozen-encoder similarity after exact matching on $K$, $T$, and target $M$, consistent with reduced lexical distinctiveness between sources. Appendix~\ref{app:similarity-matched} provides the corresponding controlled analyses.

% Our controlled analyses identify two distinct sources of difficulty: ambiguity in the ownership route and similarity between the latent sources. Figure~\ref{fig:mixing_intensity} shows that recovery difficulty is not monotonic in mixing intensity. For two sources, intermediate mixing admits many plausible ownership patterns, whereas maximal switching constrains the route to an alternating pattern and can therefore reduce structural uncertainty. The larger \((K,T)\) conditions remain substantially harder, although these cross-curve differences should not be interpreted as an isolated source-count effect because \(K\) and \(T\) vary jointly.

% Recovery also worsens as the latent sources become more similar. After exact matching on \(K\), \(T\), and target \(M\), the extreme-minus-low similarity contrast is \(+0.0524\) MP-WER (95% CI \([0.0230,0.0818]\)), consistent with reduced lexical distinctiveness between sources. Appendix~\ref{sec:difficulty_factors} provides the detailed numerical results, matching procedure, and additional ownership and boundary metrics.

\section{Conclusion and Limitations}
\label{sec:conclusion}

% We introduced text unmixing as word-level reconstruction from complete lexical evidence and a known source count.
% \method{} learns history-conditioned source ownership, uses first-appearance labels and completion-safe constraints to search for an ownership route, and reconstructs the sources from the input.
% On \benchmark{}, it achieves the lowest five-track mean \MPWER{} among task-adapted free-generation and structured baselines and remains competitive with zero-shot frontier LLMs.
% The candidate-ranking results show that valid routes can still have incorrect ownership. Extending the method to infer an unknown source count \(K\) remains future work.
We introduce \problem{}, a fine-grained de-interleaving problem in which complete lexical evidence is observed but source ownership is unknown. 
\method{} approaches this by separating history-conditioned ownership prediction from deterministic sequence reconstruction: it predicts canonical ownership routes under completion-safe constraints and then reconstructs each source directly from the indexed input occurrences. Across the five principal \benchmark{} tracks, \method{} achieves the lowest mean \MPWER{} among task-adapted free-generation and structured baselines while preserving every observed word occurrence exactly, and remains competitive with zero-shot frontier LLMs. 
These results suggest that, when lexical evidence is fully observed, decoupling ownership inference from lexical regeneration provides a reliable alternative to direct source-text regeneration.

There are several limitations remaining in this work. 
First, the current formulation assumes that the source count \(K\) is known. Jointly inferring the number of sources and their ownership is an important extension. Second, finite-beam decoding guarantees structural validity but not correct ownership, and our candidate-ranking analysis indicates that substantial ranking error remains even among legal routes. Improving route scoring or reranking may therefore yield further gains. 
Third, our formulation isolates ownership recovery under complete lexical observations. Real pipelines may additionally contain ASR/OCR insertions, deletions, or substitutions, so extending evidence-preserving routing to noisy or incomplete observations remains an open problem.

\clearpage
\section*{Ethics Statement}
We use public text and AMI/ICSI annotations and collect no new human-subject
data; consequential source-attribution claims require independent evidence and
human review.

\section*{Reproducibility Statement}
The appendix and supplement specify the methods and provide the code, tests, configurations, and commands needed to reconstruct the datasets and rerun the reported experiments. The project code will be made publicly available upon acceptance of the paper.

\section*{AI Use Statement}
Generative AI is used to improve manuscript readability and in the model-assisted first stage of the semantic-identifiability audit described in Appendix~\ref{app:semantic-identifiability}. Specifically, GPT-5.6 Sol screened 120 examples for materially different legal partitions that could remain linguistically plausible; a human reviewer then examined all 120 cases and made the final judgments. The authors reviewed all AI-assisted work and remain responsible for the content and reported conclusions.

\bibliography{references}
\bibliographystyle{iclr2027_conference}

\clearpage
\appendix
\addtocontents{toc}{\protect\setcounter{tocdepth}{2}}
\renewcommand{\contentsname}{Appendix Contents}
\tableofcontents
\clearpage
\section{Experimental Details}
\label{app:experimental-details}

\subsection{Baselines and Matched Controls}
\label{app:baseline-details}

\paragraph{Comparison baselines.}
Table~\ref{tab:baseline-implementation} summarizes the components required to
reproduce each baseline comparison.  The task-finetuned generator predicts $K$
source-word arrays in nested JSON and is trained with assistant-token causal
language-modeling loss.  The task-specific baseline routing heads use the last LM subword state
aligned to each input word; \method{} instead uses the existing vocabulary head.

\ifturnmarkerresults
The task-matched turn-marker generator
\citep{park-georgiou-2018-multimodal} uses the same backbone, training
examples, adaptation family, and generation budget as source-array generation.
Its target retains the mixed-stream order, copies every word occurrence,
and inserts \texttt{<S0>}, \texttt{<S1>}, and subsequent canonical markers
before the first word and whenever ownership changes.  The first marker is
\texttt{<S0>}, new markers follow first-appearance order, and all $K$ markers
must occur.  A strict full-output parser rejects explanations, malformed or
redundant markers, missing sources, and non-word-layer items.  The training
context is 1,152 LM tokens because two complete training targets exceed 1,024;
this avoids dropping or truncating examples.  All other data, optimization,
seed, and evaluation choices follow the matched source-generation protocol.
\fi

\begin{table*}[t]
\centering
\footnotesize
\setlength{\tabcolsep}{4pt}
\caption{Implementation summary and trainable parameter counts for the comparison
baselines.}
\label{tab:baseline-implementation}
\textit{Learning and model state}\par\smallskip
\begin{tabular}{@{}>{\raggedright\arraybackslash}p{0.17\textwidth}>{\raggedright\arraybackslash}p{0.25\textwidth}>{\raggedright\arraybackslash}p{0.37\textwidth}>{\raggedright\arraybackslash}p{0.13\textwidth}@{}}
\toprule
System & Scoring / state & Trainable head & Loss \\
\midrule
Free generation & causal source-text likelihood & none beyond QLoRA & assistant-token LM \\
\ifturnmarkerresults
Turn-marker generation & causal marked-stream likelihood & none beyond QLoRA & assistant-token LM \\
\fi
Global Transformer & bidirectional contextual emissions & 256-d projection; 2-layer, 8-head encoder; linear emission & occurrence CE \\
EMA prototype & emission plus cosine to per-source EMA ($\lambda=.75$, scale 4) & 256-d projection/query; shared biases & occurrence CE \\
RGS-CRF & global emissions plus $4\!\times\!4$ transitions & Global head plus transition matrix & exact legal-route CRF NLL \\
Legal projection & edit cost to generated source arrays; EMA tie guide & none in projection & none \\
\bottomrule
\end{tabular}

\vspace{0.7em}
\textit{Inference and trainable size}\par\smallskip
\begin{tabular}{@{}>{\raggedright\arraybackslash}p{0.24\textwidth}>{\raggedright\arraybackslash}p{0.50\textwidth}r@{}}
\toprule
System & Decoder / projection & Trainable parameters \\
\midrule
Free generation & greedy JSON generation & 64,929,792 \\
\ifturnmarkerresults
Turn-marker generation & greedy word-and-marker generation & 64,929,792 \\
\fi
Global Transformer & legal RGS beam, $B=4$ & 67,166,468 \\
EMA prototype & stateful legal RGS beam, $B=4$ & 65,652,486 \\
RGS-CRF & emission/transition RGS beam, $B=4$ & 67,166,484 \\
Legal projection & exact shuffle DP; otherwise anchor beam 512 & 0 (projection)$^\dagger$ \\
\midrule
\multicolumn{3}{@{}l}{\footnotesize $^\dagger$Projection adds no trainable
parameters to its generator and guide checkpoints.}\\
\bottomrule
\end{tabular}
\end{table*}

Legal projection first accepts a proposal when an exact shuffle-certificate DP
verifies it (state limit 250,000).  Otherwise, a beam of 512 maximizes ordered
word anchors, charges unmatched observed and skipped generated words as edit
cost, and uses the aligned EMA route only to break edit ties; every result is
then reconstructed by stable filtering.

The structured baselines draw on self-attention
\citep{vaswani-etal-2017-attention}, prototype classification
\citep{snell-etal-2017-prototypical}, conditional random fields
\citep{lafferty-etal-2001-crf}, and edit-distance projection
\citep{wagner-fischer-1974-string}. These citations identify component families;
the task-specific implementations in Table~\ref{tab:baseline-implementation}
use different ownership scores or projection objectives.

\paragraph{Matched representation control.}
PI-Route is a matched representation control for first-appearance
canonicalization, using exhaustive permutation-invariant training
\citep{yu-etal-2017-permutation}. It replaces canonical
targets with arbitrary source labels and enumerates all $K!$ global relabelings
for the minimum LM-token-normalized negative log-likelihood.  It matches \method{}
in backbone, trainable parameters, data, learning rate, epochs, context length,
source counts, training runs, beam budget, reconstruction map, and metrics.

\subsection{Training configuration}
\label{app:training-configuration}

The task-adapted comparisons use post-trained Qwen3.5-4B with NF4 quantization,
rank-32 QLoRA, $\alpha=64$, dropout 0.05, two epochs, effective batch size 16,
and three training runs \citep{dettmers-etal-2023-qlora}.  \method{} and PI-Route use
learning rate $2\times10^{-4}$ and a context of 1,024 LM tokens. The Global, EMA,
and RGS-CRF routers use $10^{-4}$ and a context of 512 LM tokens. A length audit finds
that 3.4\% of the supervised \method{} sequences exceed 512 LM tokens and none
exceeds 1,024.

The cross-backbone experiments use one training run per backbone for
Qwen/Qwen3.5-4B, Qwen/Qwen3.5-9B, and google/gemma-7b-it.  They keep the 10,000
training mixtures, two epochs, rank-32 QLoRA, a context of 1,024 LM tokens, effective
batch size 16, known-$K$ interface, and evaluation examples fixed.  Batch size
and gradient accumulation are adjusted to preserve the effective batch.  Each
backbone passes a synthetic known-route test, atomic-symbol checks, exact mask
checks, stable reconstruction, non-emptiness, and word-layer evaluation before
the reported evaluation.

\paragraph{Hardware.}
The EPOR--PI-Route runtime comparison used the same NVIDIA GeForce RTX
5090 (32 GB) host for all six training runs (three seeds per method).
Appendix~\ref{app:pi-route} specifies the timing scope.

\subsection{Prompts and output formats}
\label{app:prompts}

This section gives the prompts, output formats, and parsing rules used in the
experiments.  Input words are indexed and JSON-encoded before applying the
model-specific chat template.  The panels below distinguish system
instructions, instance templates, substituted fields, and expected responses
visually without altering the prompt text. In these verbatim prompts, input
``tokens'' denote task-level word occurrences.

\begin{center}
\setlength{\fboxsep}{0pt}\setlength{\fboxrule}{0.8pt}
\fcolorbox{PromptFrame}{white}{\begin{minipage}{0.965\linewidth}
\promptheader{EPOR prompt template}
\vspace{5pt}
\hspace{0.025\linewidth}\begin{minipage}{0.95\linewidth}
\promptband{PromptSystemBg}{PromptSystem}{System message}
\vspace{3pt}{\footnotesize\ttfamily\raggedright
You perform source routing under an external grammar.\\
Emit exactly one single-digit source label per indexed input token, with no
spaces or punctuation.\\
The decoder enforces legal labels. Do not emit any explanation.\par}
\vspace{6pt}
\promptband{PromptUserBg}{PromptUser}{User template}
\vspace{3pt}{\footnotesize\ttfamily\raggedright
The sequence contains \promptvar{K} latent sources. Source order is arbitrary.\\
Emit exactly \promptvar{T} digits, one label in [0,
\promptvar{K\_minus\_1}] per token.\\
The first label must be 0. A new label may only be introduced as one plus the
largest previous label. Every source must occur at least once.\\[3pt]
index\textbackslash ttoken\\
\promptvar{indexed\_json\_tokens}\par}
\vspace{6pt}
\promptband{PromptOutputBg}{PromptOutput}{Expected response}
\vspace{3pt}{\footnotesize\ttfamily\raggedright Compact labels:
\textcolor{PromptOutput}{\bfseries 012001\ldots}\par}
\vspace{6pt}
\end{minipage}
\end{minipage}}
\vspace{3pt}

{\small\emph{EPOR prompt template.  Color distinguishes the system
instruction, instance template, and expected response; blue text marks
substituted fields.}}
\end{center}

The displayed template is the decimal-codec prompt. Its target is the
concatenated zero-based route without separators, corresponding to the
one-based mathematical labels in Section~\ref{sec:method}. The prompt is
loss-masked; route symbols and the assistant termination suffix are supervised.
At inference, the decoder emits exactly $T$ ownership symbols, excludes EOS
from the route alphabet, and rejects unsupported output tokens.

\begin{center}
\setlength{\fboxsep}{0pt}\setlength{\fboxrule}{0.8pt}
\fcolorbox{PromptFrame}{white}{\begin{minipage}{0.965\linewidth}
\promptheader{Compact source-array generation prompt template}
\vspace{5pt}
\hspace{0.025\linewidth}\begin{minipage}{0.95\linewidth}
\promptband{PromptSystemBg}{PromptSystem}{System message}
\vspace{3pt}{\footnotesize\ttfamily\raggedright
You reconstruct latent source texts from an interleaved token sequence.\\
Return exactly one JSON array per source, where every source is represented as
an array of word tokens.\\
Source order is arbitrary. Return only one nested JSON array and no
explanation.\par}
\vspace{6pt}
\promptband{PromptUserBg}{PromptUser}{User template}
\vspace{3pt}{\footnotesize\ttfamily\raggedright
The observed sequence interleaves \promptvar{K} latent sources while preserving
word order inside each source.\\
Reconstruct \promptvar{K} source token arrays from the \promptvar{T} observed
tokens.\\[3pt]
index\textbackslash ttoken\\
\promptvar{indexed\_json\_tokens}\par}
\vspace{6pt}
\promptband{PromptOutputBg}{PromptOutput}{Expected response}
\vspace{3pt}{\footnotesize\ttfamily\raggedright JSON source token arrays:
\textcolor{PromptOutput}{\bfseries [["word",\ldots],["word",\ldots]]}\par}
\vspace{6pt}
\end{minipage}
\end{minipage}}
\vspace{3pt}

{\small\emph{Compact source-array generation prompt template.  The formatting
separates instructions supplied to the model from instance fields and
returned JSON.}}
\end{center}

The parser selects the last nested JSON value containing exactly $K$ arrays of
strings.  Parse failure is scored as an invalid or empty prediction; no legal-route repair is applied to the free-generation output.  The comparison measures
this task-finetuned generator--parser pipeline.

\paragraph{Evidence-explicit source-array generation.}
The second task-finetuned Qwen comparison and the DeepSeek-V4 and
DeepSeek-V4.1-Flash evaluations use the prompt shown below.

\begin{center}
\setlength{\fboxsep}{0pt}\setlength{\fboxrule}{0.8pt}
\fcolorbox{PromptFrame}{white}{\begin{minipage}{0.965\linewidth}
\promptheader{Evidence-explicit source-array generation prompt template}
\vspace{5pt}
\hspace{0.025\linewidth}\begin{minipage}{0.95\linewidth}
\promptband{PromptSystemBg}{PromptSystem}{System message}
\vspace{3pt}{\footnotesize\ttfamily\raggedright
You reconstruct latent source sequences by partitioning indexed observed word
occurrences. Return exactly one non-empty JSON array per source, where every
element is a word string copied from the input.\\
Use every observed word occurrence exactly once. Do not add, delete, duplicate,
correct, merge, split, or paraphrase words. Preserve increasing input-index
order within every source. Source order is arbitrary. Return only one valid
nested JSON array and no explanation.\par}
\vspace{6pt}
\promptband{PromptUserBg}{PromptUser}{User template}
\vspace{3pt}{\footnotesize\ttfamily\raggedright
The observed sequence contains \promptvar{T} indexed word occurrences from
\promptvar{K} latent sources. Partition all occurrences into exactly
\promptvar{K} non-empty source token arrays. Copy each token string exactly once
across the arrays. Do not add, delete, duplicate, correct, merge, split, or
paraphrase any token. Preserve increasing input-index order within each source.\\[3pt]
index\textbackslash ttoken\\
\promptvar{indexed\_json\_tokens}\par}
\vspace{6pt}
\promptband{PromptOutputBg}{PromptOutput}{Expected response}
\vspace{3pt}{\footnotesize\ttfamily\raggedright JSON source token arrays:
\textcolor{PromptOutput}{\bfseries [["word",\ldots],["word",\ldots]]}\par}
\vspace{6pt}
\end{minipage}
\end{minipage}}
\vspace{3pt}

{\small\emph{Evidence-explicit source-array generation prompt template used for the
 task-finetuned comparison and hosted zero-shot models.}}
\end{center}

The task-finetuned Qwen system uses the same data, adaptation budget, three
training runs, decoding budget, parser, and evaluation examples as the compact
generator. Hosted evaluations use the same evidence-explicit prompt and parser
in non-thinking mode. Calls were issued through the DeepSeek API or Alibaba
Cloud Model Studio under the provider-facing aliases available at evaluation
time. 
For all generators, invalid JSON or the wrong number of arrays is an empty
prediction, and no repair is applied.

\FloatBarrier

\section{Dataset Construction}
\label{app:data-construction}

\subsection{Corpora and lexical units}
\label{app:corpora and lexical units}

Tables~\ref{tab:data-provenance}--\ref{tab:real-serialization-specification}
summarize corpus-specific unitization, split construction, and
serialization-track statistics, respectively. The corpora are
\citep{bowman-etal-2015-large,williams-etal-2018-broad,
merity-etal-2016-wikitext,zhang-etal-2015-character,
mccowan-etal-2005-ami,
janin-etal-2003-icsi,wang-etal-2021-layoutreader}.
Repeated strings are not deduplicated inside an example; each word occurrence
retains its input position. A word occurrence is a task-level unit produced by
the corpus-specific unitizer, including punctuation where retained. LM tokens
refer to the model's subword encoding and may subdivide a word occurrence.
In Tables~\ref{tab:data-provenance} and~\ref{tab:benchmark-specification},
``Examples / groups'' lists materialized examples first and provenance groups
second; different renderings of the same group count as separate examples.

\begin{table*}[t]
\centering
\footnotesize
\caption{Corpora, roles, and preprocessing used in the experiments.}
\label{tab:data-provenance}
\resizebox{0.995\textwidth}{!}{%
\begin{tabular}{@{}>{\raggedright\arraybackslash}p{0.13\textwidth}>{\raggedright\arraybackslash}p{0.17\textwidth}>{\raggedright\arraybackslash}p{0.16\textwidth}>{\raggedright\arraybackslash}p{0.46\textwidth}@{}}
\toprule
Corpus & Role & Examples / groups & Source unit and preprocessing \\
\midrule
SNLI 1.0 & Train & 10,000 / 10,000 & Premise and hypothesis spans; deduplication across held-out splits, NFKC normalization, whitespace collapse, and deterministic regex unitization. \\
SNLI 1.0 & Development & 1,000 / 1,000 & Same preprocessing as training; source/text-disjoint from training and test data. \\
MultiNLI 1.0 & Controlled evaluation & 2,640 / 240 & Premises only; NFKC normalization, whitespace collapse, and eight whitespace units per source. Prompt and document identifiers define provenance. \\
MultiNLI 1.0 & Paired construction & 480 / 240 & Same fixed-length preprocessing; identical source word occurrences under controlled and independent-clock mixing. \\
MultiNLI 1.0 & Unequal-length stress & 540 / 180 & Variable source spans; independent-clock, long-run, and bursty mixing. Prompt and document identifiers define provenance. \\
WikiText-103 & Domain transfer & 1,448 / 1,448 & Length-filtered sentence spans materialized with the deterministic regex unitizer. \\
AG News & Domain transfer & 2,500 / 2,500 & Official-test article text filtered to 12--80 regex units. \\
AMI 1.6.2 & Speech-domain and temporal-serialization transfer & 1,000 / 59 meetings & NFKC-normalized XML word nodes, including punctuation; 64-unit windows. \\
ICSI core & Speech-domain transfer (consumed reserve) & 300 / 15 meetings & NFKC-normalized word nodes with deterministic timing repair; 64-unit windows. \\
ICSI core & Temporal serialization (development) & 400 / 15 meetings & A development view replayed from the frozen timestamp-merge construction; 64-unit windows. \\
ReadingBank & Spatial serialization & 833 / 828 documents & XML-reading-order segments retained as deterministic metadata-derived reading flows; 32- or 64-unit windows. \\
WikiText-103 & Digital serialization & 1,200 / 400 groups & The same 400 source groups rendered with smooth, bursty, and unequal-rate virtual event schedules. \\
\bottomrule
\end{tabular}%
}%
\end{table*}

\begin{table*}[t]
\centering
\scriptsize
\setlength{\tabcolsep}{2.5pt}
\caption{Core \benchmark{} specification.  All held-out groups are
provenance-disjoint from training/development; the paired and stress sets are
also disjoint from the controlled MultiNLI evaluation.}
\label{tab:benchmark-specification}
\resizebox{\textwidth}{!}{%
\begin{tabular}{lllllllll}
\toprule
Split & Corpus & Source unitization & Examples / groups & $K$ & $T$/length & Interleaving & Disjointness & Seed \\
\midrule
Train & SNLI & NFKC + regex words & 10,000 / 10,000 & 2--4 & variable & controlled IID & held-out source/text & 2027 \\
Development & SNLI & NFKC + regex words & 1,000 / 1,000 & 2--4 & variable & controlled IID & train/test source/text & 2027 \\
Controlled evaluation & MultiNLI & first 8 whitespace units/source & 2,640 / 240 & 2--4 & 16/24/32 & 11 controlled intensities & source + document & 2701 \\
Paired construction & MultiNLI & first 8 whitespace units/source & 480 / 240 & 2--4 & 16/24/32 & controlled + renewal clock & source + document & 13013 \\
Unequal-length stress & MultiNLI & variable source spans & 540 / 180 & 2--4 & variable & clock + long-run + bursty & source + document & 20270905 \\
\bottomrule
\end{tabular}}
\end{table*}

\begin{table*}[t]
\centering
\scriptsize
\setlength{\tabcolsep}{3pt}
\caption{Serialization-transfer extension. Temporal and spatial tracks use
real metadata; digital profiles simulate a concurrent-production mechanism
with virtual event-time schedules and reuse the same source groups. Counts
describe frozen evaluation views before model inference.}
\label{tab:real-serialization-specification}
\begin{tabular}{@{}llrrlll@{}}
\toprule
Track & Serialization mechanism & Examples & Groups & $K$ counts & $T$ counts & Mean $M$ \\
\midrule
AMI & annotated word-time merge & 1,000 & 59 & 2:667; 3:333 & 64:1,000 & 0.275 \\
ICSI dev & annotated/repaired word-time merge & 400 & 15 & 2:200; 3:200 & 64:400 & 0.290 \\
ReadingBank & metadata-derived reading flows & 833 & 828 & 2:300; 3:300; 4:233 & 32:383; 64:450 & 0.176 \\
Digital--smooth & virtual event-time merge & 400 & 400 & 2:134; 3:133; 4:133 & 32:201; 64:199 & 0.582 \\
Digital--bursty & virtual event-time merge & 400 & 400 & 2:134; 3:133; 4:133 & 32:201; 64:199 & 0.266 \\
Digital--unequal-rate & virtual event-time merge & 400 & 400 & 2:134; 3:133; 4:133 & 32:201; 64:199 & 0.531 \\
\bottomrule
\end{tabular}
\end{table*}

\paragraph{Operational ReadingBank sources.}
For each page, the observed sequence is the left-to-right/top-to-bottom page
serialization, indexed $0,\ldots,n-1$. ReadingBank also supplies a permutation
$q=(q_1,\ldots,q_n)$ such that selecting observed positions in that order
recovers the DOCX-XML reading-order sequence. We split $q$ immediately before
each descent, $q_{j+1}<q_j$; the resulting maximal increasing runs define the
reading flows. Each observed position receives the ID of the run that contains
it and its rank within that run. These are deterministic, metadata-derived
flows, not human-annotated semantic sources or layout regions. For example,
$q=(0,3,1,4,2,5)$ yields runs $(0,3)$, $(1,4)$, and $(2,5)$ and observed-order
labels $(0,1,2,0,1,2)$. Released sidecars make the construction traceable
through \texttt{page\_id}, \texttt{window\_start}, \texttt{window\_length},
\texttt{gold\_route}, \texttt{source\_local\_indices}, and
\texttt{decomposition\_version}; for instance,
\path{aqa-70372-sow-humang-globalsgoverng.docx:5} uses the 32-word window
starting at observed position 112. Its first twelve observed positions have
flow labels $(0,0,0,0,1,1,2,2,2,3,3,3)$; the first six corresponding
source-local ranks are $(23,24,25,26,32,33)$ in the frozen sidecar.

Candidate windows have length 32 or 64 and stride 8 (including the final
possible start). We retain windows with $K\in\{2,3,4\}$, at least eight words
per active flow, at least $K$ switches, and a valid one-to-one normalized word
inventory. One eligible window per page is chosen by a result-blind hash with
seed 20260915; a second deterministic ranking caps each $(K,T)$ stratum at
150 examples. Documents used by the development view are excluded. This
procedure yields 833 examples from 828 test documents.

The serialization extension contains 3,433 evaluation views. Its
1,200 digital views are
three schedules over 400 shared source groups, so group-level inference keeps
the schedules together. The ReadingBank view contains 833 examples from 828
test documents and excludes all documents used to construct its development
view. Each track was filtered without model predictions and validated for
non-empty sources, one-time occurrence use, source-local order, deterministic
replay, and ownership-field isolation. We manually inspected ten examples per
serialization mechanism to verify that the mixtures followed the intended
serialization patterns.

The 10,000-example training set draws $K=(2,3,4)$ with probabilities
$(0.6,0.3,0.1)$.  The realized counts are 5,955, 2,998, and 1,047, with
25,092 source spans consumed without reuse.  Source lengths vary in training.
The controlled MultiNLI evaluation contains 240 source/document-disjoint groups,
balanced over $K$ and four bands of frozen-encoder source similarity.  Each
group is reused only across eleven target mixing intensities.  The paired
construction evaluation uses another 240 disjoint groups, with identical
source word occurrences under both route generators.  Source similarity $S$ is the mean pairwise cosine
after attention-masked mean pooling and L2 normalization with
\texttt{sentence-transformers/all-MiniLM-L6-v2}.  Empirical quartile cut points
(within $K$) are $(0.0079,0.0659,0.1332)$ for $K=2$,
$(0.0316,0.0674,0.1066)$ for $K=3$, and $(0.0476,0.0731,0.1029)$ for $K=4$.

\subsection{Interleaving mechanisms}

Let $N_{\rm switch}=\sum_{t=2}^{T}\mathbf 1[z_t\neq z_{t-1}]$.  For $T>K$,
mixing intensity is
\begin{equation}
M=\frac{N_{\rm switch}-(K-1)}{T-K}\in[0,1].
\end{equation}
The controlled mixer sets the target switch count to
\begin{equation}
N_{\rm switch}^{*}=\operatorname{round}\{(K-1)+M(T-K)\}.
\end{equation}
It allocates positive runs across sources, arranges run labels without equal
neighbors, divides each source into balanced positive run lengths, and consumes
each source from left to right.  Integer switch counts may realize the nearest
rather than the requested $M$.
The normalized measure is undefined for $T=K$. With unequal source lengths,
length imbalance can prevent a switch at every position, so the largest
attainable intensity may be below one.

The independent-clock mixer is implemented separately.  For each source $k$,
it draws i.i.d. inter-arrivals
$\Delta_{k,j}\sim\operatorname{Exponential}(1)$ and cumulative times
$\tau_{k,j}=\sum_{r\leq j}\Delta_{k,r}$.  Events are merged by
$(\tau_{k,j},k,j)$, which preserves source-local order and provides a
deterministic tie break.

\paragraph{Digital scheduling profiles.}
The digital track replays each of 400 WikiText source groups under three
virtual-time schedules. In \emph{smooth}, every source has independent
inter-arrivals $\Delta\sim\operatorname{Exponential}(1)$. In
\emph{unequal-rate}, source indices $0,1,2,3$ use rates
$(1.0,1.5,0.7,2.0)$, truncated to the realized $K$. In \emph{bursty}, the
first word of a burst follows a long-pause increment
$\operatorname{Exponential}(0.22)$, the burst length is sampled uniformly
from $\{2,3,4,5\}$ words, and subsequent words use
$\operatorname{Exponential}(4.0)$ increments. Per-source increments are
accumulated, and all word events are sorted by
$(\text{event time},\text{source index},\text{source-local index})$. The
global construction seed is 20260915; a deterministic group seed is derived
for each source group and reused across profiles. The frozen implementation is
identified by builder \texttt{digital\_virtual\_time\_v1} and serializer
\texttt{virtual\_event\_time\_merge\_v1}. The same lexical sources are used
for all three profiles, so only the scheduling mechanism changes.

The unequal-length stress test adds two mechanisms.  The long-run generator
creates a small number of extended contiguous runs, while the bursty generator
alternates short bursts with variable length.  The 180 source groups include
more than one total length for every $K\in\{2,3,4\}$.  Every builder verifies
that each source is non-empty, every indexed word occurrence is used once, source-local
order is preserved, and repeated generation with the same input state is
deterministic.

\subsection{Speech-derived text windows}

AMI words are read from speaker-specific XML and sorted by start time, end
time, speaker identifier, and source-local index.  ICSI uses an annotated start
when available, otherwise an annotated end or deterministic channel-local
interpolation; wholly untimed channels are excluded.  These rules flatten
overlapping annotations rather than remove them. Both builders scan 64-unit
windows at stride 32, retain $K\in\{2,3\}$, require at least eight words per
represented speaker, and supply only the flattened words and $K$ to the model.
The domain-transfer analysis uses the frozen 1,000-window AMI view from 59
meetings and a 300-window ICSI reserve from 15 reserve meetings. The
temporal-serialization comparison reuses the same AMI examples and predictions,
so the two AMI roles are not independent evaluations. For ICSI, it instead uses
a separate frozen 400-window development view from 15 validation meetings. The
ICSI development and reserve meeting sets are disjoint, and no example is
shared between them. Intervals resample complete meetings.

\subsection{Semantic-identifiability audit}
\label{app:semantic-identifiability}

We inspected a stratified panel of 120 controlled-MultiNLI examples spanning
$K\in\{2,3,4\}$, four source-similarity bands, and five mixing settings. The
GPT-5.6 Sol reviewer agent using the extra-high reasoning setting was shown the
mixed text, $K$, and the recorded source partition and asked to construct a
materially different legal partition that remained linguistically plausible.
The screen found no clear alternative for 103 examples, marked seven as
uncertain, and flagged three as plausible alternatives and seven as strongly
underdetermined. A human reviewer then examined all 120 examples with access to
the mixed text, $K$, and recorded source partition, using semantic, grammatical,
and discourse coherence. The human reviewer found no materially different
plausible alternative for the 103 examples initially marked clear. Of the 17
non-clear cases, only two $K=4$ examples, at target $M=0.75$ and $M=1.0$,
retained a plausible alternative; the other 15 were judged to have one
defensible partition. Thus, the two-stage model-assisted and human audit found
no materially different plausible alternative in 118 examples, while two
retained one. The stratified panel was designed to probe difficult cases, so
this descriptive rate is neither an unbiased prevalence estimate nor a human
ceiling on \MPWER{}.

\FloatBarrier

\section{Formal Details and Guarantees}
\label{app:epor-implementation}

\subsection{Valid Output Space and Permutation Invariance}
\label{app:valid-output-space}

Equation~\ref{eq:labeled-assignment-space} defines all surjective labeled
source assignments. The symmetric group $S_K$ acts on this space by global
relabeling, $(\pi\!\cdot\!z)_t=\pi(z_t)$. Because every label occurs, each
assignment orbit has $K!$ members. Stable filtering maps globally relabeled
assignments to source tuples that differ only by the same permutation.
Therefore Equation~\ref{eq:valid-output-space} is a quotient output space: it
contains non-empty, occurrence-conserving, order-preserving source tuples
modulo global source permutation. This support follows from the task definition
and does not depend on a learned score or a particular prediction interface.
Repeated words can make different indexed assignments lexically
indistinguishable (Appendix~\ref{app:alignment-dp}), and different partitions
can also remain semantically plausible.

\subsection{Evaluation Metrics}
\label{app:additional-metrics}

For predicted sources $\widehat Y$ and gold sources $Y$, \MPWER{} is
\begin{equation}
\operatorname{MP\mbox{-}WER}(\widehat Y,Y)=
\min_{\pi}\frac{\sum_{k=1}^{K'}
d_{\mathrm{edit}}(\widehat y_k,y_{\pi(k)})}
{\max(1,\sum_{y\in Y}|y|)},
\end{equation}
where $K'=\max(|\widehat Y|,|Y|)$ and the smaller side is padded with
empty sources. For a predicted ownership route, ownership accuracy is
\begin{equation}
\operatorname{OwnAcc}(\widehat z,z^\star)
=\max_{\pi\in S_K}\frac{1}{T}\sum_{t=1}^{T}
\mathbf 1[\pi(\widehat z_t)=z_t^\star].
\end{equation}

Ownership accuracy and macro F1 use the source permutation that maximizes
positional agreement. This
alignment remains necessary with canonical labels: different predicted and
gold partitions can introduce corresponding sources in different orders.
Switch F1 compares the label-invariant boundary events $z_t\neq z_{t-1}$.
Source-array free-generation baselines do not provide an indexed ownership route, so their
ownership metrics are left undefined.

For turn-marker generation, we first align the generated mixed-order words to
the input by minimum word edit distance, with unit insertion, deletion, and
substitution costs. Among optimal paths, we maximize exact word matches, then
trace forward with a fixed preference for matching, substitution, deletion,
and insertion. This lexical alignment uses no ownership labels. Only exact
word matches transfer marker labels to observed positions; deleted or
substituted input words remain unassigned and count as errors. Inserted words
receive no input position. Ownership accuracy uses all $T$ input positions,
with source-label permutation optimized after word alignment. A predicted
switch requires matched labels on both adjacent input positions; every gold
switch remains in the recall denominator. Strict parsing failures receive
zero ownership scores. \MPWER{} and Evidence use the generated source texts.
Thus, for turn-marker generation, ownership accuracy is an end-to-end measure
over input positions that penalizes unmatched words; it is not a pure
attribution score conditioned on successful lexical reproduction.

Word-multiset preservation (Evidence) is one for an example if every word
string has the same total multiplicity in the reconstructed sources and the
input, and zero otherwise. The reported mean is the fraction of examples
satisfying this test; it is not a fraction of individually preserved words.
Evidence alone checks neither ownership nor source-internal order and cannot
distinguish repeated occurrences with the same surface form. Task-level
structural validity requires $K$ non-empty sources, exact occurrence use, and
source-local order. For assignment-based systems, route validity additionally
checks assignment length, label range, canonical first appearance, and use of
every source. Recovery and support metrics are reported separately.

\subsection{Repeated-Word Alignment}
\label{app:alignment-dp}

Repeated surface forms can admit several indexed alignments even when the gold
source strings are fixed.  A dynamic-programming state
$c=(c_1,\ldots,c_K)$ records the consumed prefix length of every source.  At
input position $t$, transition $k$ is allowed exactly when
$c_k<|y_k|$ and $x_t=y_{k,c_k+1}$.  Counts of paths reaching the same state are summed, and the
terminal count is the number of indexed-source alignment paths.  Source indices
remain fixed inside this computation.  Thus $x=(a,a)$ with two identical
singleton sources has two indexed paths but only one unlabeled partition.  We
use this count only as an ambiguity diagnostic; \MPWER{} operates on the
reconstructed source strings.

\subsection{Canonical Representation and Supervision}
\label{app:support-proofs}

For any partition of the $T$ input positions into $K$ non-empty unlabeled
blocks, order the blocks by their smallest position and label them
$1,\ldots,K$. The resulting route satisfies Equation~\ref{eq:legal-rgs}.
Conversely, forgetting the labels of a route in $\mathcal Z_{T,K}$ recovers
the indexed partition. These maps are inverse. First-appearance
canonicalization therefore selects one member from each orbit of the task-level
assignment space, giving
\begin{equation}
\mathcal P^{\mathrm{lab}}_{T,K}/S_K\cong\mathcal Z_{T,K}.
\end{equation}
There are $K!S(T,K)$ labeled surjections and $S(T,K)$ canonical routes, where
$S(T,K)$ is the Stirling number of the second kind. This correspondence is
between canonical routes and indexed partitions, not between routes and unique
source strings. Canonicalization removes label redundancy but does not resolve
semantic ambiguity.

\paragraph{Ownership encoding scheme and notation.}
Each word occurrence is assigned one output LM token, while its indexed input
representation may contain several LM tokens. The encoding scheme checks that ownership
symbols have distinct token IDs and encode atomically both alone and in
adjacent sequences. Decimal symbols are used when these checks pass; otherwise,
verified existing reserved tokens provide the route alphabet. No vocabulary
entries or ownership head are added. Unsupported or non-atomic encodings fail
validation. Mathematical routes use labels $1,\ldots,K$; the implementation
and decimal prompts use $\widetilde z_t=z_t-1\in\{0,\ldots,K-1\}$.
This offset changes notation, not the partition or admissibility rule.

\paragraph{Supervision implementation.}
The indexed input and $K$ are formatted with the model-specific chat template,
followed by the serialized canonical route and the standard assistant
termination suffix. Prompt and padding positions are excluded from the loss;
route and termination positions are supervised. Algorithm~\ref{alg:epor-training}
and Equation~\ref{eq:route-objective} specify the training procedure, and
Appendix~\ref{app:training-configuration} gives the QLoRA configuration.

\subsection{Completion-Safe Decoding}
\label{app:completion-safe-proof}

Consider a canonical prefix before position $t$, with largest label
$m_{t-1}$ and $m_0=0$. The label range in
Equation~\ref{eq:legal-action-set} is necessary and sufficient for preserving
first appearance in the next step. After choosing $a$, there are
$K-\max(m_{t-1},a)$ unseen sources and $T-t$ remaining positions. A completion
exists if and only if the latter count is at least the former. Each missing
source can be introduced once in order, after which surplus positions can be
assigned to an existing source. Thus $\mathcal A_t$ contains exactly the
actions that admit a complete canonical route. Starting from the empty prefix
and applying this rule for $T$ steps yields a member of $\mathcal Z_{T,K}$.

\paragraph{Implementation details.}
Algorithm~\ref{alg:epor-inference} gives the complete decoding procedure. The
admissibility mask is computed separately for each retained prefix and keeps
the original score of every legal extension:
\[
\ell_t(a\mid z_{<t})=
\begin{cases}
\log p_\theta(a\mid x,K,z_{<t}), & a\in\mathcal A_t(z_{<t}),\\
-\infty, & \text{otherwise}.
\end{cases}
\]
This mask restricts the candidate set rather than redefining a normalized
probability distribution over legal labels. EOS is excluded from the route
alphabet, and decoding stops after exactly $T$ ownership symbols. An empty beam
or a route that fails the encoding, canonicality, non-emptiness, order, or
evidence-preservation checks is rejected rather than repaired. Finite-beam
search approximates maximization of Equation~\ref{eq:epor-score}; it does not enumerate the full
support or guarantee a globally optimal route. Every accepted route is
nevertheless a complete canonical partition. Unlike training, inference uses
predicted prefixes and fixed-length termination; the supervised assistant
suffix is not used to decide when to stop routing.

\subsection{Exact Reconstruction Guarantee}
\label{app:reconstruction-guarantee}

The admissibility rule characterizes the full feasible route space, whereas a
finite beam generally explores only part of it. Beam pruning cannot make a
retained prefix invalid, but it can discard the globally highest-scoring
route. Validity therefore holds for every positive beam width without an
optimality guarantee. For a complete route, the sets $\{t:z_t=k\}$ are
disjoint, cover $\{1,\ldots,T\}$, and are non-empty. Reading each set in
increasing index uses every input position once and preserves order within each
assigned source. It therefore reproduces the materialized input-occurrence
multiset without assuming correct ownership. These guarantees concern the
task-level occurrences, not raw-file formatting. Unit tests enumerate routes
for small $(T,K)$ and compare Equation~\ref{eq:legal-action-set} with exhaustive
completions.

\subsection{Training and Inference Algorithms}
\label{app:algorithms}

Algorithms~\ref{alg:epor-training} and~\ref{alg:epor-inference} give the
complete training and inference procedures described in Section~\ref{sec:method}.

\begin{algorithm}[H]
\captionsetup[algorithm]{font=footnotesize}
\caption{\method{} training}
\label{alg:epor-training}
\begin{algorithmic}[1]
\scriptsize
\Require Examples $\{(x_i,K_i,z_i^\star)\}$; causal LM $p_\theta$
\State Canonicalize each $z_i^\star$ by first appearance
\ForAll{minibatches $\mathcal M$}
  \ForAll{$(x,K,z^\star)\in\mathcal M$}
    \State $P\gets\Call{FormatPrompt}{x,K}$
    \State $v^\star\gets\Call{SerializeRoute}{z^\star}$
    \State Append assistant termination suffix to $v^\star$
    \State Form teacher-forced input $[P;v^\star]$
    \State Mask prompt and padding from loss
  \EndFor
  \State Compute $\mathcal L_{\mathrm{route}}$ on $\mathcal M$ using Equation~\ref{eq:route-objective}
  \State Update QLoRA parameters
\EndFor
\State \Return adapted scorer $p_\theta$
\end{algorithmic}
\end{algorithm}

\begin{algorithm}[H]
\caption{\method{} completion-safe inference}
\label{alg:epor-inference}
\footnotesize
\begin{algorithmic}[1]
\scriptsize
\Require $x=(x_1,\ldots,x_T)$; $K$; scorer $p_\theta$; beam width $B$
\State $\mathcal B_0\gets\{(\epsilon,0)\}$
\For{$t=1,\ldots,T$}
  \State $\mathcal C\gets\varnothing$
  \ForAll{$(z_{<t},s)\in\mathcal B_{t-1}$}
    \State $q_t(a)\gets\log p_\theta(a\mid x,K,z_{<t})$
    \Statex \hspace{\algorithmicindent}\hspace{\algorithmicindent}for ownership symbols $a$
    \State Compute $\mathcal A_t(z_{<t})$ by Equation~\ref{eq:legal-action-set}
    \ForAll{$a\in\mathcal A_t(z_{<t})$}
      \State Add $(z_{<t}\mathbin{\|}a,s+q_t(a))$ to $\mathcal C$
    \EndFor
  \EndFor
  \State $\mathcal B_t\gets\Call{TopB}{\mathcal C,B}$
\EndFor
\State $\widehat z_B\gets\arg\max_{(z,s)\in\mathcal B_T}s$
\State \Return $\widehat z_B,\ R(x,\widehat z_B)$
\end{algorithmic}
\end{algorithm}

\section{Extended Experiments}
\label{app:more-experiments}

\subsection{Ownership-interface ablations and diagnostics}
\label{app:ownership-controls}

The adaptation-by-support comparison uses one common Qwen3.5-4B checkpoint,
indexed EPOR prompt, canonical decimal codec, known $K$, evaluation set, and
deterministic reconstruction.  Adapted rows use the three task-adapted QLoRA
adapters; frozen rows use no adapter or demonstrations.  Constrained rows use
the same completable-prefix mask, beam size four, and exactly $T$ generated
symbols as EPOR.  Unmasked rows also request exactly $T$ symbols, but the full
vocabulary remains available. This removes the entire legality mask, not only
the remaining-capacity condition.  Unsupported symbols (including an unexpected
EOS), noncanonical routes, missing sources, and wrong-length outputs are scored
as emitted and are never repaired.  Unit tests verify length preservation,
illegal-symbol scoring, canonical-route checks, EOS handling, and deterministic
evidence reconstruction.  In the reported unmasked runs, every generated
position nevertheless used an in-codec ownership symbol, so stable filtering
consumed every input occurrence and evidence preservation remained 1.0.  Their
incomplete-structure scores instead come from routes that omit one or more of
the known $K$ sources.  An unsupported symbol would be retained as invalid and
would reduce evidence preservation if its position could not be assigned to an
emitted source bucket.

Table~\ref{tab:mechanism-search-contract} makes the matched and unmatched
search support explicit for these four adaptation-by-support conditions.

\begin{table*}[t]
\centering
\scriptsize
\setlength{\tabcolsep}{3pt}
\caption{Search settings for the adaptation-by-support ablations. ``LM beam''
uses the decoder's original beam-score convention; no legal-subset
renormalization is added.}
\label{tab:mechanism-search-contract}
\resizebox{\textwidth}{!}{%
\begin{tabular}{lccccc}
\toprule
Condition & Beam & Score normalization & Generated length & EOS & Codec validation \\
\midrule
Adapted, constrained (EPOR) & 4 & LM beam score & exactly $T$ & excluded as route action & atomic; reject unsupported \\
Frozen, constrained & 4 & LM beam score & exactly $T$ & excluded as route action & atomic; reject unsupported \\
Adapted, unconstrained & 4 & LM beam score & exactly $T$ & suppressed before $T$ & retain as invalid ($-1$) \\
Frozen, unconstrained & 4 & LM beam score & exactly $T$ & suppressed before $T$ & retain as invalid ($-1$) \\
\bottomrule
\end{tabular}%
}
\end{table*}

For the fixed-template intervention, the causal LM sees the canonical sequence
in implementation notation,
$0,1,\ldots,K-1,0,\ldots$ at earlier ownership positions, truncated to the
current prefix length.  This sequence is independent of both the decoded and
gold assignments.  The decoder still computes its legal actions from the
unmodified decoded route.  The graded intervention preserves first
appearances; once all $K$ labels have appeared, it deterministically rotates a
preselected fraction of model-visible prior labels.  Thus both interventions
preserve output support and reconstruction while selectively removing
assignment-specific information from the causal context.

The matched fixed-template history ablation uses the same route-independent
legal template during training and evaluation.  Gold canonical ownership
symbols remain the supervised targets, but the input side of the causal shift
contains only the mixed-text prompt and the fixed template prefix.  It
therefore learns the same LM-token-level objective without observing
example-specific $z_{<t}$.  Backbone, QLoRA rank,
learning rate, context length, examples, epochs, seeds, codec, beam budget,
legality mask, and reconstruction are identical to EPOR.  Evaluation again
shows the fixed template to the LM while the external decoder maintains the
true generated prefix solely to compute legal actions.

\subsection{Full MultiNLI metrics}

Table~\ref{tab:primary-full} supplements the main system comparison with
position-level and boundary metrics.

\begin{table}[H]
\centering
\scriptsize
\setlength{\tabcolsep}{3pt}
\caption{Full results on the controlled MultiNLI \benchmark{} test set.  \MPWER{} is
mean$\pm$sample SD over three training runs; remaining entries are three-run
means, with corresponding SDs provided in the supplementary results.}
\label{tab:primary-full}
\begin{tabular}{lcccc}
\toprule
Method & \MPWER{} $\downarrow$ & Ownership acc. $\uparrow$ & Switch F1 $\uparrow$ & Evidence $\uparrow$ \\
\midrule
Compact source-array generation & \ConfirmFreeWERMean$\pm$\ConfirmFreeWERSD & -- & -- & \ConfirmFreeEvidenceMean \\
Evidence-explicit source-array generation & \StrictFreeWERMean$\pm$\StrictFreeWERSD & -- & -- & \StrictFreeEvidenceMean \\
\ifturnmarkerresults
Turn-marker generation & 0.4271$\pm$0.0096 & 0.6991 & 0.8081 & 0.3957 \\
\fi
Global Transformer & \ConfirmGlobalWERMean$\pm$\ConfirmGlobalWERSD & \ConfirmGlobalTokenAccMean & \ConfirmGlobalSwitchFOneMean & \ConfirmGlobalEvidenceMean \\
EMA prototype & \ConfirmEmaWERMean$\pm$\ConfirmEmaWERSD & \ConfirmEmaTokenAccMean & \ConfirmEmaSwitchFOneMean & \ConfirmEmaEvidenceMean \\
RGS-CRF & \ConfirmCrfWERMean$\pm$\ConfirmCrfWERSD & \ConfirmCrfTokenAccMean & \ConfirmCrfSwitchFOneMean & \ConfirmCrfEvidenceMean \\
Legal projection & \ConfirmProjectionWERMean$\pm$\ConfirmProjectionWERSD & \ConfirmProjectionTokenAccMean & \ConfirmProjectionSwitchFOneMean & \ConfirmProjectionEvidenceMean \\
\method{} & \ConfirmDirectRgsWERMean$\pm$\ConfirmDirectRgsWERSD & \ConfirmDirectRgsTokenAccMean & \ConfirmDirectRgsSwitchFOneMean & \ConfirmDirectRgsEvidenceMean \\
\bottomrule
\end{tabular}
\end{table}

\subsection{Generation Baselines: Prompt Variants and Error Diagnostics}
\label{app:generation-diagnostics}

\begin{table}[H]
\centering
\scriptsize
\setlength{\tabcolsep}{3.6pt}
\caption{Generation-pipeline diagnostics, mean$\pm$sample SD over three runs.
Conditional metrics use each method's successfully parsed subset.}
\label{tab:generation-diagnostics}
\begin{tabular}{lcccc}
\toprule
Prompt & Parse & Limit hit & Conditional \MPWER{} & Conditional Evidence \\
\midrule
Compact & \ConfirmFreeParseSuccess$\pm$\ConfirmFreeParseSuccessSD & n/a & \ConfirmFreeConditionalWER$\pm$\ConfirmFreeConditionalWERSD & \ConfirmFreeConditionalEvidence$\pm$\ConfirmFreeConditionalEvidenceSD \\
Evidence-explicit & \StrictFreeParseSuccess$\pm$\StrictFreeParseSuccessSD & \StrictFreeLimitHit$\pm$\StrictFreeLimitHitSD & \StrictFreeConditionalWER$\pm$\StrictFreeConditionalWERSD & \StrictFreeConditionalEvidence$\pm$\StrictFreeConditionalEvidenceSD \\
\bottomrule
\end{tabular}
\end{table}

Table~\ref{tab:generation-diagnostics} separates parsing and truncation behavior
from conditional reconstruction quality. All-example evaluation, including
parse failures, is the reported comparison.  The parser
accepts only a nested JSON value with exactly $K$ arrays of strings; invalid
JSON or wrong cardinality is scored as an empty prediction, and no output is
repaired. Termination metadata were not retained for the compact-prompt runs,
so their limit-hit rate is unavailable rather than inferred.
Conditional-on-parse metrics describe different method-specific subsets and
are diagnostic rather than a replacement for the all-example comparison.  Relative
to evidence-explicit generation, \method{} changes all-example Evidence by
\EPORMinusStrictFreeEvidenceDelta{} (95\% CI
\EPORMinusStrictFreeEvidenceCI).

\ifturnmarkerresults
\FloatBarrier
\subsection{Task-matched turn-marker generation}
\label{app:turn-marker}

Table~\ref{tab:turn-marker-comparison} reports the three-run MultiNLI result.
The first difference column is \method{} minus turn-marker generation; the
second is turn-marker generation minus evidence-explicit source-array
generation.  Signed differences favor the row-named system when positive for
ownership accuracy, Switch F1, and Evidence, but when negative for
\MPWER{}.  Confidence intervals use 20,000 source-group bootstrap draws after
averaging training runs within each example.

\begin{table*}[!htbp]
\centering
\scriptsize
\setlength{\tabcolsep}{4pt}
\caption{Task-matched turn-marker generation on the 2,640-example MultiNLI
evaluation.  The turn-marker column is mean$\pm$sample SD over three training
runs.  Difference columns give paired point estimates with 95\% source-group
bootstrap intervals.}
\label{tab:turn-marker-comparison}
\resizebox{\textwidth}{!}{%
\begin{tabular}{lccc}
\toprule
Metric & Turn-marker generation & \method{} $-$ Turn-marker & Turn-marker $-$ source-array generation \\
\midrule
\MPWER{} $\downarrow$ & 0.4271$\pm$0.0096 & $-0.0515$ $[-0.0636,-0.0395]$ & $-0.2476$ $[-0.2720,-0.2238]$ \\
Ownership acc. $\uparrow$ & 0.6991$\pm$0.0068 & $+0.0466$ $[0.0381,0.0554]$ & -- \\
Switch F1 $\uparrow$ & 0.8081$\pm$0.0032 & $+0.0432$ $[0.0338,0.0530]$ & -- \\
Evidence $\uparrow$ & 0.3957$\pm$0.0131 & $+0.6043$ $[0.5617,0.6461]$ & $+0.3143$ $[0.2778,0.3524]$ \\
\bottomrule
\end{tabular}
}
\end{table*}

The strict marker parser succeeds on
\TurnMarkerParseSuccessMean$\pm$\TurnMarkerParseSuccessSD{} of examples, and
only \TurnMarkerLimitHitMean$\pm$\TurnMarkerLimitHitSD{} reach the generation
limit.  Turn markers substantially improve on source-array generation by
removing output reordering, yet lexical regeneration remains imperfect.
On MultiNLI, \method{} retains advantages in reconstruction, occurrence-level
ownership, boundary recovery, and exact evidence preservation. This
system-level output-interface baseline isolates mixed-order marker
generation from source-array reordering, but is not a representation-only
comparison because it still regenerates every word.

Table~\ref{tab:turn-marker-transfer} extends the comparison to all five
tracks using the same three training seeds. Confidence intervals use 20,000
paired provenance-group bootstrap draws after averaging seeds within each
example: source groups for MultiNLI, meetings for AMI and ICSI, and documents
for ReadingBank. Digital resampling keeps the three schedules for each
source group together; its mean and sample SD are computed from the
equal-profile average within each seed. These pointwise intervals condition on
the three trained checkpoints.

\begin{table*}[!htbp]
\centering
\scriptsize
\setlength{\tabcolsep}{3pt}
\caption{\textbf{EPOR and turn-marker generation across five tracks.}
Turn-marker \MPWER{} is mean$\pm$sample SD over three runs. Paired
differences are \method{} minus turn-marker, with 95\% group-bootstrap
intervals; negative values favor \method{}. Ownership accuracy and Evidence
columns give \method{} / turn-marker means.}
\label{tab:turn-marker-transfer}
\resizebox{\textwidth}{!}{%
\begin{tabular}{@{}lccccc@{}}
\toprule
Track & \method{} MP & Turn-marker MP & Paired difference [95\% CI] & Ownership acc. & Evidence \\
\midrule
MultiNLI & 0.3757 & 0.4271$\pm$0.0096
& $-0.0515$ $[-0.0636,-0.0395]$
& 0.7457 / 0.6991 & 1.0000 / 0.3957 \\
AMI & 0.6320 & 0.6928$\pm$0.0312
& $-0.0608$ $[-0.0724,-0.0481]$
& 0.6205 / 0.4441 & 1.0000 / 0.6377 \\
ICSI development & 0.6161 & 0.6690$\pm$0.0185
& $-0.0529$ $[-0.0704,-0.0353]$
& 0.6228 / 0.4805 & 1.0000 / 0.6600 \\
ReadingBank & 0.6156 & 0.6862$\pm$0.0154
& $-0.0706$ $[-0.0859,-0.0556]$
& 0.6224 / 0.4959 & 1.0000 / 0.3325 \\
Digital & 0.6474 & 0.6449$\pm$0.0088
& $+0.0026$ $[-0.0067,0.0120]$
& 0.5912 / 0.5426 & 1.0000 / 0.7769 \\
\bottomrule
\end{tabular}
}
\end{table*}

The paired \MPWER{} intervals favor \method{} on MultiNLI, AMI, ICSI, and
ReadingBank. Digital has a small point difference in favor of turn-marker
generation, with an interval spanning zero. Across all five tracks,
\method{} has higher ownership accuracy and exact evidence preservation,
whereas turn-marker generation sometimes changes the observed word multiset.
Table~\ref{tab:turn-marker-ownership} gives the ownership intervals and
boundary scores after the lexical alignment in
Appendix~\ref{app:additional-metrics}. Boundary recovery favors turn-marker
generation on AMI, ICSI, and Digital, and \method{} on MultiNLI and ReadingBank.

\begin{table}[H]
\centering
\scriptsize
\setlength{\tabcolsep}{4pt}
\caption{\textbf{Ownership and boundary recovery after lexical alignment.}
Differences are \method{} minus turn-marker, with paired 95\% group-bootstrap
intervals. Positive differences favor \method{}. All examples, including
strict parsing failures, contribute to the reported scores.}
\label{tab:turn-marker-ownership}
\resizebox{\textwidth}{!}{%
\begin{tabular}{@{}lccc@{}}
\toprule
Track & Ownership difference [95\% CI] & Switch F1: \method{} / turn-marker & Switch F1 difference [95\% CI] \\
\midrule
% BEGIN GENERATED TURN-MARKER OWNERSHIP ROWS
MultiNLI & $+0.0466$ [0.0381, 0.0554] & 0.8514 / 0.8081 & $+0.0432$ [0.0338, 0.0530] \\
AMI & $+0.1765$ [0.1632, 0.1891] & 0.2567 / 0.4469 & $-0.1902$ [-0.2090, -0.1714] \\
ICSI development & $+0.1423$ [0.1258, 0.1572] & 0.3018 / 0.5168 & $-0.2150$ [-0.2379, -0.1885] \\
ReadingBank & $+0.1264$ [0.1121, 0.1412] & 0.4459 / 0.4161 & $+0.0298$ [0.0148, 0.0449] \\
Digital & $+0.0486$ [0.0406, 0.0568] & 0.6764 / 0.7136 & $-0.0372$ [-0.0500, -0.0246] \\
% END GENERATED TURN-MARKER OWNERSHIP ROWS
\bottomrule
\end{tabular}}
\end{table}

Repeated words can admit multiple optimal lexical alignments. Reversing the
fixed tie preference changes each track's mean ownership accuracy by less
than 0.0006 and mean switch F1 by less than 0.0003. The comparisons are
unchanged under this alternative convention.

\FloatBarrier
\fi

\subsection{Frontier zero-shot language models}
\label{app:hosted-controls}

Table~\ref{tab:hosted-provenance} records the executed provider configuration
for every zero-shot row in Table~\ref{tab:confirmation}. The requested model
alias, provider route, thinking setting, temperature, and output
budget are reported for each scope.
All calls used the same evidence-explicit prompt and parser.

\begin{table*}[t]
\centering
\scriptsize
\setlength{\tabcolsep}{2.5pt}
\caption{Hosted-model execution configuration for Table~\ref{tab:confirmation}.
``Scope'' identifies the cells supplied by each run: MNLI, AMI, ICSI-dev,
ReadingBank (RB), or the three-profile digital aggregate (DIG).}
\label{tab:hosted-provenance}
\resizebox{\textwidth}{!}{%
\begin{tabular}{@{}lllllll@{}}
\toprule
Table row / scope & Provider & Requested model ID  & Thinking & Temp. & Completion-token budget \\
\midrule
Qwen3.8-Flash / all & Alibaba Cloud Model Studio & \texttt{qwen3.8-flash} & off & 0 & $\max(8192,64T)$ \\
Qwen3.8-Max / all & Alibaba Cloud Model Studio & \texttt{qwen3.8-max}  & off & 0 & $\max(8192,64T)$ \\
DeepSeek-V4-Flash / MNLI & DeepSeek API & \texttt{deepseek-v4-flash}  & off & 0 & $\max(256,16T)$ \\
DeepSeek-V4-Flash / AMI, ICSI-dev, RB, DIG & Alibaba Cloud Model Studio & \texttt{deepseek-v4-flash}  & off & 0 & $\max(8192,64T)$ \\
DeepSeek-V4-Pro / MNLI & DeepSeek API & \texttt{deepseek-v4-pro} & off & 0 & $\max(256,16T)$ \\
DeepSeek-V4-Pro / AMI, ICSI-dev, RB, DIG & Alibaba Cloud Model Studio & \texttt{deepseek-v4-pro} & off & 0 & $\max(8192,64T)$ \\
DeepSeek-V4.1-Flash / AMI, ICSI-dev, DIG & DeepSeek API & \texttt{deepseek-flash} & off & 0 & $\max(8192,64T)$ \\
DeepSeek-V4.1-Flash / MNLI, RB & DeepSeek API & \texttt{deepseek-flash}  & off & 0 & $\max(8192,64T)$ \\
\bottomrule
\end{tabular}%
}
\end{table*}

The two Table~\ref{tab:confirmation} rows marked ``mixed APIs'' combine a
locked MultiNLI DeepSeek-API control with later Alibaba Cloud transfer calls.
They therefore report the executed cells under the same requested model name,
while retaining their provider and budget differences in the Table 14.
DeepSeek-V4.1-Flash was requested as \texttt{deepseek-flash}.

\begin{table}[t]
\centering
\small
\setlength{\tabcolsep}{4pt}
\caption{MultiNLI compliance diagnostics for four hosted evaluations under
the generous $\max(8192,64T)$ budget. Metrics include all examples. Parse
failures and provider refusals receive no repair.}
\label{tab:hosted-diagnostics}
\begin{tabular}{@{}lrrrr@{}}
\toprule
System & Temp. & JSON parse $\uparrow$ & Limit hit $\downarrow$ & Refusal $\downarrow$ \\
\midrule
DeepSeek-V4.1-Flash & 0 & \HostedVFourOneMnliParse & \HostedVFourOneMnliLimit & \HostedVFourOneMnliRefusal \\
DeepSeek-V4-Pro & 0 & \HostedVFourProMnliParse & \HostedVFourProMnliLimit & \HostedVFourProMnliRefusal \\
Qwen3.8-Flash & 0 & \HostedQwenThreeEightMnliParse & \HostedQwenThreeEightMnliLimit & \HostedQwenThreeEightMnliRefusal \\
Qwen3.8-Max & 0 & \HostedQwenThreeEightMaxMnliParse & \HostedQwenThreeEightMaxMnliLimit & \HostedQwenThreeEightMaxMnliRefusal \\
\bottomrule
\end{tabular}
\end{table}

The four evaluations in Table~\ref{tab:hosted-diagnostics} use the same strict
evidence-explicit prompt, controlled-MultiNLI examples, parser, non-thinking mode,
temperature, and generous output budget. Unresolved API errors are zero after
retries; format failures remain measured failures. The DeepSeek-V4-Pro entry is
the later single-provider full-budget rerun and is distinct from the mixed-API
V4-Pro row in Table~\ref{tab:confirmation}. Ownership accuracy is undefined
for free-text source arrays. Transfer-view summaries and hashes are provided in
the supplementary experiment package.

\subsection{Ownership adaptation and history}
\label{app:mechanism-controls}

The adapted conditions in Table~\ref{tab:mechanism-full} average three models;
the frozen conditions use the same unadapted checkpoint.  All conditions use the
same batch-one evaluation setting as the MultiNLI \method{} row.

The matched route-independent fixed-template history ablation averages three
runs and differs from EPOR only in the ownership prefix visible to the causal
LM during training and evaluation.  The fixed-template-minus-EPOR differences are
\MatchedFixedTemplateMinusEporWERDelta{} \MPWER{} (95\% CI
\MatchedFixedTemplateMinusEporWERCI),
\MatchedFixedTemplateMinusEporTokenAccuracyDelta{} ownership accuracy
(\MatchedFixedTemplateMinusEporTokenAccuracyCI), and
\MatchedFixedTemplateMinusEporSwitchFOneDelta{} switch F1
(\MatchedFixedTemplateMinusEporSwitchFOneCI).  Evidence preservation and complete
$K$-way support remain 1.0 in every run.  This matched training-and-evaluation comparison tests assignment-specific
causal history while holding the decoder and reconstruction fixed.

\begin{table*}[t]
\centering
\scriptsize
\setlength{\tabcolsep}{2.5pt}
\caption{\textbf{Inference-time ownership-history interventions on MultiNLI.}
The checkpoint is fixed; only the prefix shown to the scorer changes. The
controller uses the actual candidate prefix, leaving output legality intact.}
\label{tab:history-controls}
\begin{tabular}{lccccc}
\toprule
Condition & \MPWER{} $\downarrow$ & Ownership acc. $\uparrow$ & Switch F1 $\uparrow$ & Evidence $\uparrow$ & Complete $K$-way $\uparrow$ \\
\midrule
Unmodified history & \HistoryEporReferenceWER{} & \HistoryEporTokenAccuracy{} & \HistoryEporSwitchFOne{} & \HistoryEporEvidence{} & \HistoryEporCompleteStructure{} \\
Route-independent template & \HistoryFixedTemplateInferenceWER{} & \HistoryFixedTemplateInferenceTokenAccuracy{} & \HistoryFixedTemplateInferenceSwitchFOne{} & \HistoryFixedTemplateInferenceEvidence{} & \HistoryFixedTemplateInferenceCompleteStructure{} \\
25\% eligible-label rotation & \HistoryCorruptTwentyFiveWER{} & \HistoryCorruptTwentyFiveTokenAccuracy{} & \HistoryCorruptTwentyFiveSwitchFOne{} & \HistoryCorruptTwentyFiveEvidence{} & \HistoryCorruptTwentyFiveCompleteStructure{} \\
50\% eligible-label rotation & \HistoryCorruptFiftyWER{} & \HistoryCorruptFiftyTokenAccuracy{} & \HistoryCorruptFiftySwitchFOne{} & \HistoryCorruptFiftyEvidence{} & \HistoryCorruptFiftyCompleteStructure{} \\
100\% eligible-label rotation & \HistoryCorruptHundredWER{} & \HistoryCorruptHundredTokenAccuracy{} & \HistoryCorruptHundredSwitchFOne{} & \HistoryCorruptHundredEvidence{} & \HistoryCorruptHundredCompleteStructure{} \\
\bottomrule
\end{tabular}
\end{table*}

\begin{figure}[t]
  \centering
  \includegraphics[width=0.72\linewidth]{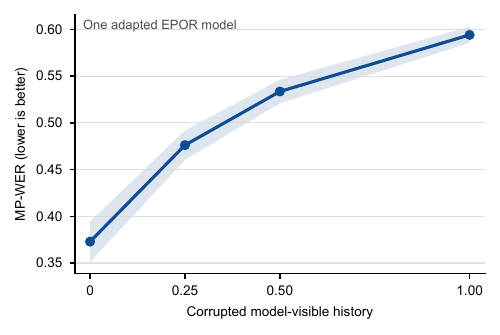}
  \caption{\textbf{Sensitivity to model-visible ownership history.} Points
  report one adapted model; bands are 95\% source-group bootstrap intervals
  over 240 groups, retaining all eleven interleavings within each group.
  Corruption begins only after all $K$ labels have appeared; the decoder's true
  legal prefix is unchanged.}
  \label{fig:history-corruption}
\end{figure}

Table~\ref{tab:history-controls} reports the numerical outcomes of the
inference-time interventions, while Figure~\ref{fig:history-corruption} shows
that recovery degrades as more assignment-specific history is corrupted. For
the model used in this curve, \MPWER{} rises from
\HistoryEporReferenceWER{} without corruption to \HistoryCorruptHundredWER{}
under full eligible-label rotation. Evidence and complete $K$-way support
remain exact, so the change is semantic rather than a legality failure. The
curve uses one trained checkpoint and an inference-time intervention, which
demonstrates sensitivity, not a uniquely identified memory mechanism.

\subsection{Permutation-invariant representation comparison}
\label{app:pi-route}

\begin{table}[t]
\centering
\small
\setlength{\tabcolsep}{4.5pt}
\caption{Matched representation comparison.  Recovery values are three-run
means; both models have \PIAuditTrainableParameters{} trainable parameters.}
\label{tab:pi-route-main}
\begin{tabular}{lccc}
\toprule
Representation & \MPWER{} $\downarrow$ & \shortstack{LM train\\sequences} & \shortstack{Rel. train\\time $\downarrow$} \\
\midrule
Canonical (\method{}) & \PIAuditDirectWERMean & \PIAuditDirectSequenceEvals{} & \textbf{1.00$\times$} \\
Arbitrary labels (PI-Route) & \PIAuditPIWERMean & \PIAuditPISequenceEvals{} & \PIAuditWallTimeRatio$\times$ \\
\bottomrule
\end{tabular}
\end{table}

\begin{table*}[h]
\centering
\footnotesize
\setlength{\tabcolsep}{4pt}
\caption{Full matched representation comparison on MultiNLI, mean$\pm$sample
SD over three training runs.}
\label{tab:pi-route}
\begin{tabular}{lccccc}
\toprule
Method & \MPWER{} $\downarrow$ & Ownership acc. $\uparrow$ & Switch F1 $\uparrow$ & Evidence $\uparrow$ & Structure $\uparrow$ \\
\midrule
\method{} & \PIAuditDirectWERMean$\pm$\PIAuditDirectWERSD & \PIAuditDirectTokenAccMean$\pm$\PIAuditDirectTokenAccSD & \PIAuditDirectSwitchFOneMean$\pm$\PIAuditDirectSwitchFOneSD & \PIAuditDirectEvidenceMean & \PIAuditDirectStructuralMean \\
PI-Route & \PIAuditPIWERMean$\pm$\PIAuditPIWERSD & \PIAuditPITokenAccMean$\pm$\PIAuditPITokenAccSD & \PIAuditPISwitchFOneMean$\pm$\PIAuditPISwitchFOneSD & \PIAuditPIEvidenceMean & \PIAuditPIStructuralMean \\
\bottomrule
\end{tabular}
\end{table*}

Tables~\ref{tab:pi-route-main} and~\ref{tab:pi-route} report the compact cost
comparison and the full recovery metrics, respectively. All paired recovery
intervals span zero.  PI-Route has the same
\PIAuditTrainableParameters{} trainable parameters but requires
\PIAuditWallTimeRatio{}$\times$ the mean training wall time.  The measured
\PIAuditPISequenceEvals{} versus \PIAuditDirectSequenceEvals{} LM sequence
evaluations (\PIAuditSequenceEvalRatio{}$\times$) count the implemented two-pass
training procedure: all permuted routes are scored without gradients, then the
chunks containing the minimum-loss routes are replayed for backpropagation.
This counter therefore includes gradient-replay rows and is distinct from the
single-pass combinatorial average $\sum_i K_i!/N$.  All six training runs used
the same NVIDIA RTX 5090 host and two-epoch, 10,000-example training loop.  Mean
recorded training time was 7,005.13 seconds for EPOR and 28,565.23 seconds for
PI-Route; thus EPOR used 75.5\% less measured wall time in this implementation.
The timing covers model training (including permutation scoring and gradient
replay for PI-Route), not evaluation, model acquisition, or data preparation.

\subsection{Difficulty factors}
\label{app:similarity-matched}

\subsubsection{Mixing intensity}
Figure~\ref{fig:mixing-full-grid} shows that recovery difficulty is not
monotonic in mixing intensity. For $(K,T)=(2,16)$, \MPWER{}
rises from 0.053 at $M=0$ to 0.249 at $M=0.4$, then falls to 0.094 at
$M=1$. With two sources, switching at every position restricts the route to an
alternating pattern, so maximal switching can reduce structural uncertainty.
The larger joint conditions remain substantially harder: at $M=1$, \MPWER{}
is 0.448 for $(3,24)$ and 0.634 for $(4,32)$. Because $K$ and $T$ vary
together, cross-curve differences describe joint task conditions rather than
an isolated source-count effect.

\subsubsection{Source similarity}

Table~\ref{tab:similarity-matched} compares the four frozen-encoder
source-similarity bands after exact matching on $K$, $T$, and target $M$.
Each $(K,T,M)$ cell receives equal weight.  Intervals use 20,000 grouped
bootstrap draws over the 240 source groups while retaining all mixing views
and averaging the three trained models within each example.

\begin{table}[H]
\centering
\scriptsize
\setlength{\tabcolsep}{3.2pt}
\caption{\textbf{Recovery across matched source-similarity strata.} Entries
are matched means with 95\% source-group bootstrap intervals. Lower \MPWER{}
is better; higher values are better otherwise.}
\label{tab:similarity-matched}
\begin{tabular}{@{}lccc@{}}
\toprule
Similarity band & \MPWER{} $\downarrow$ & Ownership acc. $\uparrow$ & Switch F1 $\uparrow$ \\
\midrule
Low & \SimilarityLowWER{} \SimilarityLowWERCi{} & \SimilarityLowTokenAcc{} \SimilarityLowTokenAccCi{} & \SimilarityLowSwitchFOne{} \SimilarityLowSwitchFOneCi{} \\
Medium & \SimilarityMediumWER{} \SimilarityMediumWERCi{} & \SimilarityMediumTokenAcc{} \SimilarityMediumTokenAccCi{} & \SimilarityMediumSwitchFOne{} \SimilarityMediumSwitchFOneCi{} \\
High & \SimilarityHighWER{} \SimilarityHighWERCi{} & \SimilarityHighTokenAcc{} \SimilarityHighTokenAccCi{} & \SimilarityHighSwitchFOne{} \SimilarityHighSwitchFOneCi{} \\
Extreme & \SimilarityExtremeWER{} \SimilarityExtremeWERCi{} & \SimilarityExtremeTokenAcc{} \SimilarityExtremeTokenAccCi{} & \SimilarityExtremeSwitchFOne{} \SimilarityExtremeSwitchFOneCi{} \\
\bottomrule
\end{tabular}
\end{table}

The extreme-minus-low contrasts are
\SimilarityExtremeMinusLowWER{} \MPWER{} (95\% CI
\SimilarityExtremeMinusLowWERCi),
\SimilarityExtremeMinusLowTokenAcc{} ownership accuracy
(\SimilarityExtremeMinusLowTokenAccCi), and
\SimilarityExtremeMinusLowSwitchFOne{} switch F1
(\SimilarityExtremeMinusLowSwitchFOneCi).  The switch-F1 interval spans zero;
the other two contrasts resolve in the direction of harder recovery for the
extreme-similarity stratum.  The independent encoder cosine is an operational
stratification measure rather than a universal semantic scale, and this
descriptive analysis does not identify a causal similarity effect.

\FloatBarrier
\subsection{Construction and backbone details}
\label{app:construction-backbone}

Table~\ref{tab:construction-diagnostics} verifies that the clock generator
induces different route statistics while holding the source words, $K$, and
$T$ fixed.  Both renderings have mean $T=24.0$, balanced $K$, zero source-
length imbalance, mean lexical overlap 0.0251, and mean log$_2$ route-space
36.6228.

\begin{table}[H]
\centering
\footnotesize
\caption{Route statistics for paired controlled and independent-clock
interleavings.}
\label{tab:construction-diagnostics}
\begin{tabular}{lrrrrr}
\toprule
Construction & Mixing $M$ & Switches & Mean run & Run var. & Ambiguity \\
\midrule
Controlled & \GenControlledMMean & \GenControlledSwitchMean & \GenControlledRunLengthMean & \GenControlledRunLengthVarianceMean & \GenControlledAmbiguityRate \\
Independent clock & \GenClockMMean & \GenClockSwitchMean & \GenClockRunLengthMean & \GenClockRunLengthVarianceMean & \GenClockAmbiguityRate \\
\bottomrule
\end{tabular}
\end{table}

\begin{table}[H]
\centering
\scriptsize
\setlength{\tabcolsep}{3.8pt}
\caption{Full cross-backbone comparison on 240 paired source groups.  Each
backbone uses one training run.  Evidence reports controlled/clock; route
methods have structural validity 1.0 in both conditions.}
\label{tab:construction-backbone}
\begin{tabular}{llccc}
\toprule
Backbone & Output & Controlled \MPWER{} $\downarrow$ & Clock \MPWER{} $\downarrow$ & Evidence C/I $\uparrow$ \\
\midrule
% Generated from stored construction, model-family, and scale-analysis summaries.
Qwen3.5-4B & \method{} & 0.4163 & 0.5189 & 1.0000/1.0000 \\
 & RGS-CRF & 0.5344 & 0.6075 & 1.0000/1.0000 \\
 & Free generation & 0.6012 & 0.6480 & 0.1708/0.1792 \\
Qwen3.5-9B & \method{} & 0.3625 & 0.4459 & 1.0000/1.0000 \\
 & Free generation & 0.5368 & 0.5886 & 0.2125/0.2333 \\
Gemma-7B-IT & \method{} & 0.4630 & 0.5488 & 1.0000/1.0000 \\
 & Free generation & 0.6449 & 0.6832 & 0.0583/0.0417 \\
\bottomrule

\end{tabular}
\end{table}

Table~\ref{tab:construction-backbone} shows the ownership-over-generation
ordering for all three evaluated backbones. For Qwen3.5-4B, \method{} minus
free generation is
\GenQwenClockDirectMinusBaselineWER{} \MPWER{} on independent clocks (95\% CI
\GenQwenClockDirectMinusBaselineWERCI).  The Gemma-7B-IT difference is
\GenGemmaClockDirectMinusBaselineWER{}
(\GenGemmaClockDirectMinusBaselineWERCI), and the Qwen3.5-9B difference is
\GenQwenNineClockDirectMinusBaselineWER{}
(\GenQwenNineClockDirectMinusBaselineWERCI).  Within \method{}, 9B minus 4B is
\GenQwenDirectScaleControlledWER{} on controlled data
(\GenQwenDirectScaleControlledWERCI) and \GenQwenDirectScaleClockWER{} on
clocks (\GenQwenDirectScaleClockWERCI).  The 9B model yields lower \method{}
\MPWER{} on both constructions.

\begin{table}[H]
\centering
\scriptsize
\setlength{\tabcolsep}{3.5pt}
\caption{Unequal-length construction comparison on 180 paired source groups.
$^{\dagger}$The final row receives gold source lengths and is a privileged
reference rather than a deployable baseline.}
\label{tab:shortcut-stress}
\begin{tabular}{lcccc}
\toprule
Method & Clock \MPWER{} & Long-run \MPWER{} & Bursty \MPWER{} & Min. evidence $\uparrow$ \\
\midrule
\method{} & 0.500 & 0.288 & 0.469 & 1.000 \\
RGS-CRF & 0.568 & 0.347 & 0.612 & 1.000 \\
Task-finetuned source-array generation & 0.623 & 0.658 & 0.592 & 0.139 \\
Random legal route & 0.726 & 0.775 & 0.697 & 1.000 \\
Equal contiguous blocks & 0.683 & 0.278 & 0.759 & 1.000 \\
Gold-length contiguous blocks$^{\dagger}$ & 0.668 & 0.000 & 0.740 & 1.000 \\
\bottomrule
\end{tabular}
\end{table}

Table~\ref{tab:shortcut-stress} shows that \method{} has the lowest \MPWER{}
among the task-trained systems under all three unequal-length constructions.
The equal-block heuristic is slightly better on the long-run construction
(0.278 versus 0.288), while its substantially larger error on clock and bursty
routes exposes the construction-specific shortcut.

\subsection{Domain transfer}
\label{sec:cross-domain-direct}

Without target-domain tuning, \method{} lowers \MPWER{} relative to RGS-CRF on
WikiText, the shared AMI view, and the ICSI reserve but increases it on AG News,
indicating non-uniform domain transfer.

\begin{table}[H]
\centering
\footnotesize
\setlength{\tabcolsep}{3pt}
\caption{Domain transfer, mean$\pm$SD over three training runs. WikiText and
AG News intervals resample examples; AMI and ICSI-reserve intervals resample meetings.
AMI is the same frozen 1,000-window view used in the temporal-serialization
analysis; ICSI here denotes the separate 300-window consumed reserve.
$\Delta$ is \method{} minus RGS-CRF.}
\label{tab:cross-domain-direct}
\begin{tabular}{lrrr}
\toprule
Domain & RGS-CRF \MPWER{} & \method{} \MPWER{} & $\Delta$ \MPWER{} [95\% CI] \\
\midrule
WikiText & \CrossWikiCrfWERMean$\pm$\CrossWikiCrfWERSD & \CrossWikiDirectWERMean$\pm$\CrossWikiDirectWERSD & \CrossWikiDirectMinusCrfWER{} \CrossWikiDirectMinusCrfWERCI \\
AG News & \CrossAgCrfWERMean$\pm$\CrossAgCrfWERSD & \CrossAgDirectWERMean$\pm$\CrossAgDirectWERSD & \CrossAgDirectMinusCrfWER{} \CrossAgDirectMinusCrfWERCI \\
AMI & \CrossAmiCrfWERMean$\pm$\CrossAmiCrfWERSD & \CrossAmiDirectWERMean$\pm$\CrossAmiDirectWERSD & \CrossAmiDirectMinusCrfWER{} \CrossAmiDirectMinusCrfWERCI \\
ICSI reserve & \CrossIcsiCrfWERMean$\pm$\CrossIcsiCrfWERSD & \CrossIcsiDirectWERMean$\pm$\CrossIcsiDirectWERSD & \CrossIcsiDirectMinusCrfWER{} \CrossIcsiDirectMinusCrfWERCI \\
\bottomrule
\end{tabular}
\end{table}

\ifrealserializationresults
\subsection{Serialization transfer}
\label{app:real-serialization-results}

Table~\ref{tab:serialization-per-view} decomposes the source-array and
structured-baseline serialization aggregates
in Table~\ref{tab:confirmation}; the turn-marker comparison is in
Table~\ref{tab:turn-marker-transfer}. The digital aggregate is the equal-profile
mean over smooth, bursty, and unequal-rate schedules applied to the same 400
source groups. The ICSI row is the 400-window development view, not the
300-window reserve in Table~\ref{tab:cross-domain-direct}.

\begin{table*}[t]
\centering
\scriptsize
\setlength{\tabcolsep}{2.8pt}
\caption{Source-array and structured baselines by serialization view. Entries are mean \MPWER{} over three
training seeds. ``Free-c'' and ``Free-e'' denote compact and evidence-explicit
task-finetuned generation. $\Delta$ compares \method{} with the lowest-mean
source-array or structured baseline shown in that view; 95\% CIs resample provenance
groups after averaging seeds. Negative differences favor \method{}. Best and
second-best means are bold and underlined.}
\label{tab:serialization-per-view}
\resizebox{\textwidth}{!}{%
\begin{tabular}{@{}lrrrrrrrl@{}}
\toprule
View & Free-c & Free-e & Global & EMA & RGS-CRF & Projection & \method{} & $\Delta$ to best alternative [95\% CI] \\
\midrule
AMI & 0.7275 & 0.7418 & \underline{0.6406} & 0.6596 & 0.6483 & 0.6431 & \textbf{0.6320} & $-0.0086$ [$-0.0175$, $+0.0003$] \\
ICSI development & 0.7439 & 0.7530 & \underline{0.6389} & 0.6674 & 0.6495 & 0.6483 & \textbf{0.6161} & $-0.0228$ [$-0.0377$, $-0.0078$] \\
ReadingBank & 0.8415 & 0.8547 & 0.6820 & 0.7427 & \underline{0.6555} & 0.6843 & \textbf{0.6156} & $-0.0399$ [$-0.0517$, $-0.0281$] \\
Digital--smooth & 0.7193 & 0.7283 & 0.7718 & 0.7326 & 0.7467 & \textbf{0.6741} & \underline{0.6750} & $+0.0009$ [$-0.0128$, $+0.0152$] \\
Digital--bursty & 0.7431 & 0.7307 & 0.6763 & 0.6794 & \underline{0.6406} & 0.6764 & \textbf{0.6156} & $-0.0251$ [$-0.0409$, $-0.0090$] \\
Digital--unequal-rate & 0.7306 & 0.7352 & 0.7482 & 0.7114 & 0.7223 & \underline{0.6788} & \textbf{0.6517} & $-0.0270$ [$-0.0413$, $-0.0130$] \\
\bottomrule
\end{tabular}%
}
\end{table*}

Among the systems in Table~\ref{tab:serialization-per-view},
\method{} has the lowest mean \MPWER{} on five of the six views. On the smooth
digital profile, legal projection is lower by 0.0009 and the paired interval
spans zero. Against the per-view runner-up, the intervals exclude zero for
ICSI development, ReadingBank, bursty digital, and unequal-rate digital, but
not for AMI. Hosted outcomes remain in Table~\ref{tab:confirmation}, with
compliance diagnostics in Table~\ref{tab:hosted-diagnostics}.
\fi

\subsection{Candidate quality and ranking}
\label{app:beam-search}

\begin{figure*}[t]
  \centering
  \includegraphics[width=0.46\textwidth,trim=0bp 0bp 335bp 0bp,clip]{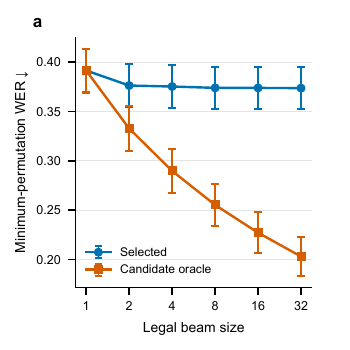}\hfill
  \includegraphics[width=0.46\textwidth,trim=168bp 0bp 183bp 0bp,clip]{direct_rgs_beam_support.pdf}

  \vspace{4pt}

  \includegraphics[width=0.46\textwidth,trim=320bp 0bp 0bp 0bp,clip]{direct_rgs_beam_support.pdf}
  \caption{\textbf{Candidate quality, selection, and cost.}  Selected and
  candidate-oracle \MPWER{} are shown across legal beam sizes, together with
  gold-route coverage, runtime, and peak allocated memory.  Points average
  three trained models; intervals resample source groups.}
  \label{fig:beam-support}
\end{figure*}

Figure~\ref{fig:beam-support} separates candidate generation from candidate
ranking.  At beam 32, the best candidate under gold \MPWER{} reaches
\MechanismBeamThirtyTwoOracleWER{}, whereas the model score selects
\MechanismBeamThirtyTwoSelectedWER{}, a regret of
\MechanismBeamThirtyTwoWERRegret{}.  The search can therefore retain candidates with substantially lower
reconstruction error than the route selected by the learned score. This gap
is measured within the finite candidate set; it is not a bound on the error
of exact score maximization. The oracle uses the reference sources and is
not deployable. All candidates satisfy the same structural constraints,
regardless of beam width. A reference-free reranker over complete $(x,z)$
candidates is a possible extension; the reported selector is unchanged.

\section{Extended Related Work}
\label{app:broader_connections}

Word-level text unmixing is also related to several established problem families outside the closest NLP settings reviewed in Section~\ref{sec:related-work}. These connections concern different aspects of the problem: formal-language work studies order-preserving interleavings of discrete sequences, document understanding studies recovery of reading structure from spatial and visual evidence, and blind source separation studies recovery of latent sources from observed mixtures. 
They help situate text unmixing among broader source-recovery and de-interleaving problems.

\subsection{Shuffle Languages and String Unshuffling}
\label{app:shuffle}

Order-preserving sequence interleaving has a classical formulation in formal-language theory through the \emph{shuffle product}. For strings \(u\) and \(v\), their shuffle contains the strings obtainable by interleaving symbols from \(u\) and \(v\) while preserving the internal order of each string. Early work studied the corresponding \emph{merge recognition} problem: given an observed string and candidate component strings, determine whether the former is an order-preserving merge of the latter \citep{mansfield1982merge}. For any fixed number of given component strings, merge recognition admits a polynomial-time algorithm, whereas allowing the number of components to vary as part of the input makes the problem NP-complete \citep{mansfield1983complexity}. The shuffle product has subsequently been studied more broadly as an operation in formal-language theory and combinatorics on words, including its expressive properties and connections to models of concurrent processes \citep{berstel2010shuffle,restivo2015shuffle}.

A related line of work considers \emph{shuffle squares}, where the component string itself is unknown but the two components are constrained to be identical. Given a string \(w\), the problem asks whether there exists some \(u\) such that \(w \in u \shuffle u\). Recognizing such shuffle squares is NP-complete \citep{rizzi2013shuffle,buss2014unshuffling}.

These problems share with word-level text unmixing the combinatorial structure of order-preserving interleaving, but differ in what is observed and what must be inferred. Classical merge recognition is supplied with the candidate component strings, while shuffle-square recognition replaces this information with the strong constraint that the two latent components are identical. In word-level text unmixing, by contrast, all \(K\) source sequences are latent, need not be identical, and have exchangeable source identities. The observation preserves the individual word occurrences but not their source ownership, so recovery requires inferring an occurrence-level partition from the lexical stream. The classical complexity results above therefore do not directly establish the computational complexity of word-level text unmixing; rather, shuffle theory provides a formal analogue of its underlying order-preserving interleaving structure.

\subsection{Reading-Order Recovery in Document Understanding}
\label{app:reading_order}

Document analysis provides another closely related setting in which serialization can obscure latent textual structure. Reading-order detection has long been treated as part of recovering the logical organization of complex documents. Earlier work developed representations and evaluation procedures for reading order over document regions, including layouts with ordered, unordered, and grouped relations \citep{clausner2013reading}. More recent document-understanding systems learn reading structure using textual features together with explicit spatial or visual information. LayoutReader predicts reading order from text and layout representations \citep{wang-etal-2021-layoutreader}, while ERNIE-Layout incorporates reading-order prediction into multimodal pretraining over textual, spatial, and visual features \citep{peng2022ernielayout}. Other approaches directly address incorrect document serialization through token-path prediction \citep{zhang-etal-2023-reading} or model reading order as ordering relations among layout elements rather than as a single global permutation \citep{zhang2024modeling}.

The spatial track of UNMIXBENCH is motivated by the same broad failure mode: multiple coherent reading flows can be serialized into a single stream in an order that does not preserve those flows. The information interface is different, however. For the ReadingBank track, document metadata are used during benchmark construction to derive deterministic operational reading flows and the corresponding ownership routes, whereas the evaluated model receives only the resulting lexical stream \(x\) and source count \(K\). Layout coordinates, page geometry, visual features, and the reading-order metadata used for construction are not supplied to the model.

The spatial text-unmixing track should therefore not be interpreted as an alternative to layout-aware reading-order prediction. Reading-order systems aim to exploit surviving document geometry and multimodal structure to determine how page elements should be traversed. Text unmixing instead studies a downstream recovery setting in which such attribution information has already been discarded while the lexical occurrences themselves remain available. It asks whether the latent reading flows can still be recovered from the serialized lexical evidence alone.

\subsection{Conceptual Relation to Blind Source Separation}
\label{app:bss}

A broader conceptual connection arises with blind source separation (BSS), which studies recovery of latent source signals from observed mixtures without direct access to the individual sources. Early adaptive methods sought to separate unknown independent sources from sensor observations \citep{jutten1991blind}. Independent component analysis subsequently provided a statistical formulation based on recovering components with reduced statistical dependence \citep{comon1994ica}, while information-theoretic approaches demonstrated blind separation by maximizing information transfer \citep{bell1995infomax}. Classical surveys and algorithmic treatments further developed the statistical principles and practical methods underlying BSS and independent component analysis \citep{cardoso1998blind,hyvarinen2000ica}.

The relation to word-level text unmixing is conceptual rather than algebraic. Classical BSS generally observes superpositions or transformations of simultaneously active source signals and recovers the underlying signals under assumptions about the sources and mixing process, often including statistical independence. Word-level text unmixing does not assume an additive mixture, a linear mixing matrix, or an ICA-style independence model. Instead, the mixing mechanism is serialization: discrete word occurrences from several sources are interleaved into a shared sequence while each occurrence itself remains intact. What is lost is the correspondence between an occurrence and its latent source.

Thus, text unmixing is better viewed as a discrete source-attribution problem under order-preserving interleaving than as an instance of classical BSS. The broader connection is that both settings separate evidence produced by latent sources after source identity has become entangled in the observation process. In the text-unmixing setting studied here, this abstraction is specialized to complete lexical evidence with missing ownership.

\end{document}